\documentclass[journal,12pt,onecolumn]{IEEEtran}
\usepackage{cite}
\usepackage{amsmath,amssymb,amsfonts}
\usepackage{algorithmic}
\usepackage{graphicx}
\usepackage{textcomp}
\usepackage{cite}
\usepackage{amsmath,amssymb,amsfonts}
\usepackage{algorithm,algorithmic}
\usepackage{hyperref}
\hypersetup{hidelinks=true}
\usepackage{caption}
\usepackage{subcaption}
\usepackage{dblfloatfix}
\usepackage{stmaryrd}
\usepackage{setspace}

\def\BibTeX{{\rm B\kern-.05em{\sc i\kern-.025em b}\kern-.08em
		T\kern-.1667em\lower.7ex\hbox{E}\kern-.125emX}}
\makeatletter
\let\save@mathaccent\mathaccent
\newcommand*\if@single[3]{%
	\setbox0\hbox{${\mathaccent"0362{#1}}^H$}%
	\setbox2\hbox{${\mathaccent"0362{\kern0pt#1}}^H$}%
	\ifdim\ht0=\ht2 #3\else #2\fi
}
\newcommand*\rel@kern[1]{\kern#1\dimexpr\macc@kerna}
\newcommand*\widebar[1]{\@ifnextchar^{{\wide@bar{#1}{0}}}{\wide@bar{#1}{1}}}
\newcommand*\wide@bar[2]{\if@single{#1}{\wide@bar@{#1}{#2}{1}}{\wide@bar@{#1}{#2}{2}}}
\newcommand*\wide@bar@[3]{%
	\begingroup
	\def\mathaccent##1##2{%
		\let\mathaccent\save@mathaccent
		\if#32 \let\macc@nucleus\first@char \fi
		\setbox\z@\hbox{$\macc@style{\macc@nucleus}_{}$}%
		\setbox\tw@\hbox{$\macc@style{\macc@nucleus}{}_{}$}%
		\dimen@\wd\tw@
		\advance\dimen@-\wd\z@
		\divide\dimen@ 3
		\@tempdima\wd\tw@
		\advance\@tempdima-\scriptspace
		\divide\@tempdima 10
		\advance\dimen@-\@tempdima
		\ifdim\dimen@>\z@ \dimen@0pt\fi
		\rel@kern{0.6}\kern-\dimen@
		\if#31
		\overline{\rel@kern{-0.6}\kern\dimen@\macc@nucleus\rel@kern{0.4}\kern\dimen@}%
		\advance\dimen@0.4\dimexpr\macc@kerna
		\let\final@kern#2%
		\ifdim\dimen@<\z@ \let\final@kern1\fi
		\if\final@kern1 \kern-\dimen@\fi
		\else
		\overline{\rel@kern{-0.6}\kern\dimen@#1}%
		\fi
	}%
	\macc@depth\@ne
	\let\math@bgroup\@empty \let\math@egroup\macc@set@skewchar
	\mathsurround\z@ \frozen@everymath{\mathgroup\macc@group\relax}%
	\macc@set@skewchar\relax
	\let\mathaccentV\macc@nested@a
	\if#31
	\macc@nested@a\relax111{#1}%
	\else
	\def\gobble@till@marker##1\endmarker{}%
	\futurelet\first@char\gobble@till@marker#1\endmarker
	\ifcat\noexpand\first@char A\else
	\def\first@char{}%
	\fi
	\macc@nested@a\relax111{\first@char}%
	\fi
	\endgroup
}
\makeatother
\renewcommand{\bar}{\widebar}

\newcommand{\probP}{\text{I\kern-0.15em P}}
\newcommand{\expE}{\text{I\kern-0.15em E}}
\newcommand{\dynamicA}{\ensuremath{\mathit{A}_*}}
\newcommand{\estdynamicA}{\ensuremath{\widehat{A_T}}}
\newcommand{\covmatrixV}{\ensuremath{\mathit{V}}}
\newcommand{\controlB}{\ensuremath{\mathit{B}_*}}
\newcommand{\WeinerP}{\ensuremath{\mathit{\mathbb{W}}}}
\newcommand{\scaledstate}{\ensuremath{\mathit{Y}}}
\newcommand{\WeinerPP}{\ensuremath{\mathit{\mathbb{W}}}}
\newcommand{\WeinerControl}{\ensuremath{\mathit{\mathbb{U}}}}
\newcommand{\weinerP}{\ensuremath{\mathit{\mathbb{V}}}}
\newcommand{\actionU}{\ensuremath{\mathit{U}}}
\newcommand{\eigenvalue}{\ensuremath{\lambda}}
\newcommand{\realeigenvalue}{\ensuremath{\Re({\lambda})}}
\newcommand{\state}{\ensuremath{\mathit{X}}}

\newcommand{\dimstate}{\ensuremath{p}}
\newcommand{\dimaction}{\ensuremath{q}}
\newcommand{\dimnoise}{\ensuremath{r}}

\newtheorem{theorem}{Theorem}

\newtheorem{assumption}{Assumption}
\newtheorem{proposition}{Proposition}
\newtheorem{remark}{Remark}
\newtheorem{lemma}{Lemma}

\makeatletter
\def\@tvsp{\mathchoice{{}\mkern-4.5mu}{{}\mkern-4.5mu}{{}\mkern-2.5mu}{}}
\def\ltrivert{\left|\@tvsp\left|\@tvsp\left|}
\def\rtrivert{\right|\@tvsp\right|\@tvsp\right|}
\makeatother
\newcommand\tnorm[1]{\ltrivert#1\rtrivert}
\makeatletter
\def\@tvsp{\mathchoice{{}\mkern-4.5mu}{{}\mkern-4.5mu}{{}\mkern-2.5mu}{}}
\def\ltrivert{\left|\@tvsp\left|\@tvsp\left|}
\def\rtrivert{\right|\@tvsp\right|\@tvsp\right|}
\makeatother
\begin{document}
	\title{~\\On the Effect of Instability on Learning
		Continuous-Time Linear Control Systems}
	\author{Reza Sadeghi Hafshejani\IEEEmembership{}, Mohamad Kazem Shirani Faradonbeh \IEEEmembership{}
		\thanks{}
		\thanks{The authors are with the Department of Mathematics, Southern Methodist University, Dallas, TX 75205 USA (e-mail: rezas@smu.edu).}
	}
	
	\maketitle
	
	\begin{abstract}
		We study the problem of system identification for stochastic continuous-time dynamics, based on a single finite-length state trajectory. We present a method for estimating the possibly unstable open-loop matrix by employing properly randomized control inputs. Then, we establish theoretical performance guarantees showing that the estimation error decays with trajectory length, a measure of excitability, and the signal-to-noise ratio, while it grows with dimension. Numerical illustrations that showcase the rates of learning the dynamics, will be provided as well. To perform the theoretical analysis, we develop new technical tools that are of independent interest. That includes non-asymptotic stochastic bounds for highly non-stationary martingales and generalized laws of iterated logarithms, among others.
	\end{abstract}
	

	%
	\IEEEpeerreviewmaketitle

	
	\section{Introduction}
	
	Stochastic linear dynamical systems are powerful models for capturing the evolutions of random environments. They appear frequently in various areas such as portfolio optimization \cite{wang2020continuous}, algorithmic trading \cite{cartea2018algorithmic}, and quantitative biology \cite{bailo2018optimal}. In many applications, there are non-negligible uncertainties about the system dynamics, necessitating it to be learned by interacting with the environment~\cite{caines1984adaptive,rusnak1995optimal}. Practically speaking, the above learning of the system dynamics can use the data of a single state trajectory, which is presumably unstable in the sense that the state variables suffer from big fluctuations and have increasing magnitudes over time. In the sequel, we elaborate both the causes and consequences of such instabilities, as well as their effects on system identification.

	Unstable linear dynamical systems inevitably arise in engineering and biology \cite{simorgh2020system,anglart2011nuclear,grizzi2006cancer,rajinikanth2010identification}, among others. For example, (nuclear) reactors that are preferred to operate stably for long periods of time, can become unstable due to structural breaks or adversarial attacks \cite{simorgh2020system,anglart2011nuclear}. So, the issue needs to be identified and addressed according to a \emph{shortest-possible} state trajectory~\cite{simorgh2020system,anglart2011nuclear}. In microbiology, a tumor can be viewed as a dynamical system, where the metastatic cancer cells are associated with loss of stability~\cite{grizzi2006cancer}. Therefore, accurate estimation of system dynamics becomes crucial for understanding the main causes of the disease. Furthermore, in such systems, the identification needs to be performed based on \emph{a single} state trajectory, since the underlying patient dynamics or the reactor breaks, both are highly likely to be idiosyncratic~\cite{davis2019dynamical,simorgh2020system}. 
	
	The problem of system identification is conventionally studied for stable discrete-time dynamics. Both asymptotic~\cite{lai1983asymptotic,ljung1998system,faradonbeh2019applications} and non-asymptotic results on the estimation errors are established for a single system  \cite{zheng2020non,oymak2019non}, or a family of multiple related systems~\cite{modi2024joint}. On the other hand, the latest findings on unstable system identification indicate that learning under instability can be essentially different in terms of technical challenges for establishing performance guarantees, the rates at which the estimation error decays, and the minimal assumptions for consistency~\cite{faradonbeh2018finite, simchowitz2018learning,sarkar2019near,modi2022big}.
	
	For continuous-time stochastic systems, the literature is significantly sparser. Early works establish bounds on the estimation error after infinite-time interactions with the system \cite{mandl1988consistency,duncan1992least,duncan1999adaptive,doya2000reinforcement}. More recent studies, extend these asymptotic results on estimation errors in both online \cite{faradonbeh2023online} and offline settings \cite{wang2018exploration}. To the authors' knowledge, there are only two papers on non-asymptotic learning of unstable continuous-time systems \cite{basei2022logarithmic,faradonbeh2022bayesian}. The former builds its estimation upon multiple independent system trajectories, while the latter adopts a Bayesian approach without providing a comprehensive analysis of sample complexities. That is, there are currently no theoretical results for \textit{finite-time} identification of an unstable system, especially when it comes to fully specifying the effects of all the model parameters on the quality of estimation. Filling this gap is adopted as the focus of this work. We present a comprehensive study that includes the effects of all the model parameters on the estimation error, without any assumptions on the eigen-structure of the continuous-time dynamics matrix. 
	
	From the control theory viewpoint, an accurate estimation of the underlying unstable system is necessary before that any attempts at employing a control policy can be made \cite{caines1984adaptive,rusnak1995optimal}. Therefore, having finite-time error bounds with theoretical guarantees is vital for any reliable interaction with dynamical systems. More precisely, a stable linear system can operate for extended periods of time without any issue (e.g., state explosion), clearly because the stability precludes excessively large state vectors. However, when a continuous-time linear system is unstable (in the sense of open-loop matrix having eigenvalues of positive real-parts), the state variables grow exponentially fast as time proceeds \cite{lauvdal1997stabilization}. This makes guaranteed estimation significantly more difficult, and raises the need for development of new methods for input design, as well as new technical tools for performance analysis. 
	
	We establish finite-time error bounds for estimating the dynamics matrix based on a single realization of the system trajectory. We show how the estimation error scales with time, state dimension, magnitude of the control action, a measure of excitability, and finally the eigen-structure of the underlying open-loop dynamics matrix. Our analysis shows that when the system is \emph{(marginally) stable}, the squared estimation error decays linearly with time (apart from a logarithmic factor). In \emph{unstable} systems, estimation of unexcited dynamics can be inconsistent no matter how long the trajectory grows. To address that, we apply moderately large random control inputs to ensure signal-strength in all directions of the state space, and theoretically prove that it leads to accurate estimates. In all cases, we show that the estimation error increases with the inverse of signal-to-noise ratio (SNR) and with the square-root of dimension. 
	
	To establish non-asymptotic error bounds in unstable continuous-time systems, several technical challenges must be addressed. First, instabilities (as well as marginal stabilities) induce very strong statistical dependence between every two samples from the state trajectory. Note that when the system is stable, such dependencies are weak and temporary, as the stable dynamics makes them diminish fast as time proceeds. However, unstable dynamics exacerbate the influence of every state vector on the future samples. In other words, unstable systems posses an amplifying long-term memory. Accordingly, the conventional probabilistic techniques (e.g., concentration inequalities)~\cite{basei2022logarithmic,dean2020sample} fail to effectively work, necessitating development of new techniques. Furthermore, another widely-used approach for uniformly bounding the stochastic signals via union bounds is inapplicable due to having an uncountable index set. Finally, whenever the system has both (marginally) stable and unstable modes, the strongest and weakest signals grow at very different rates with time. Specifically, since the former (latter) grows exponentially (linearly) with time, this rapid disparity between signal modes quickly renders the identification ill-conditioned~\cite{lai1982least,faradonbeh2018finite}. 
	
	In order to overcome these challenges, we establish useful representations of estimation error in terms of the sample covariance matrix of the state trajectory. Then, we prove tight deterministic upper- and lower-bounds for that matrix. Further, we generalize the existing results to matrix-valued self-normalized continuous-time martingales. The technical novelties also include uniformly bounding state magnitudes, and are obtained by leveraging several tools such as uniform bounds for Wiener processes, rescaled time indices~\cite{revuz2013continuous}, specifying the effect of eigenvalues and eigenspaces of the true open-loop dynamics, and a finite-time law of iterated logarithm~\cite{howard2021time}.  The authors expect these technicalities to be of broader interests, and provide further details in the proofs of the main and auxiliary results. 

	This paper is organized as follows. In section \ref{formulation} we formulate the problem, while Section \ref{Design} contains design of the input and estimator. Section \ref{theoretical results} presents the main results, and is followed by the proof and some intermediate results in Section~\ref{auxiliary}. Finally, Section \ref{numerical} presents numerical illustrations. Due to space constraints, technical proofs and further auxiliary lemmas are delegated to the supplementary materials\cite{hafshejani2024learning}.
	
	\paragraph*{Notation and Terminology}
	For a matrix $A$, $A^\top$ denotes its transpose and $\lambda_{\max}(A)$ denotes the real part of the eigenvalue of $A$ with the largest real part. Matrix $A$ is said to be stable if $\lambda_{\max}(A) < 0$, marginally stable if $\lambda_{\max}(A)=0$, and unstable otherwise. For an integer $n$, $\mathbb{I}_n$ denotes the identity matrix of dimension $n$. For a positive semidefinite matrix $P$, its Moore-Penrose generalized inverse is denoted as $P^\dagger$, and we use $\lambda_{\max}(P)$ and $\lambda_{\min}(P)$ to refer to the largest and smallest eigenvalues. For two symmetric matrices $P$ and $Q$, we write $P\preceq Q$ whenever $Q-P$ is positive semidefinite. For a vector $v \in \mathbb{R}^p$, $\lVert v \rVert_q$ is the $\ell_q$ norm of $v$. Finally, we define the matrix operator norm $\tnorm{A}_{q}= \sup_{v \neq 0}\lVert Av \rVert_q/\lVert v \rVert_q$. 
	


	\section{Problem Formulation}
	\label{formulation}
	We consider a linear control system, whose evolution is given by the Ito stochastic differential equation
	\begin{equation}
		\label{Linear SDE}
		d\state_t = (\dynamicA \state_t+\controlB \actionU_t )dt + Cd\WeinerP_t,
	\end{equation}
	starting from an arbitrary initial state $\state_0 \in \mathbb{R}^\dimstate$. The matrices $\dynamicA \in \mathbb{R}^{\dimstate \times \dimstate}$, and $C \in \mathbb{R}^{\dimstate \times \dimnoise}$ are the unknown open-loop dynamics matrix and the unknown noise coefficients matrix, respectively, while the input matrix $\controlB \in \mathbb{R}^{\dimstate \times \dimaction}$ is assumed to be known. Moreover, $\WeinerP_t$ is a  $\dimnoise$-dimensional Wiener process on the probability space $\left(\Omega, \mathcal{F}, \mathbb{P}\right)$. That is, $\Omega$ is the sample space, $\mathcal{F}=\left\{\mathcal{F}_t\right\}_{t \in [0,\infty]}$ is a family of non-decreasing sigma-algebras with respect to which all the random processes are measurable, and $\mathbb{P}$ is the probability measure. Further, for any positive integer  $n$, a Wiener process $\WeinerP_t$ of dimension $n$ satisfies (i) $\WeinerP_0 = 0_n$, almost surely, and (ii) $\WeinerP_t$ has independent multivariate Gaussian increments. That is, for $t_2>t_1\geq0$, the displacement $\WeinerP_{t_2}-\WeinerP_{t_1}$ is independent of $\mathcal{F}_{t_1}$, and $\WeinerP_{t_2}-\WeinerP_{t_1}\sim \mathbb{N}\left(0_n, \left(t_2-t_1\right)\mathbb{I}_{n}\right)$. 
	
	The goal is to show that through $\textit{proper design}$ of the input $\actionU_t$, we can prove $\textit{finite-time error bounds}$ for recovering the unknown dynamics matrix $\dynamicA$ from a single state trajectory. To proceed, we express the following assumption.
	
	\begin{assumption}
		We assume that $\controlB\controlB^\top$ is positive definite, and denote its smallest eigenvalue by $c = \lambda_{\min}(\controlB\controlB^\top)$.
	\end{assumption}

	\begin{remark}
		The above assumption is adopted in the existing literature \cite{liu2013robust}, \cite{caines2019stochastic}. It is a stronger notion of controllability (i.e., positivedefiniteness of controllability Gramian \cite{bertsekas2012dynamic}), that we adopt for the ease of presentation in quantifying the effects of all involved parameters on the estimation error. However, relaxing this assumption to controllability is a matter of technicality, for which the details are left to a future work.  
	\end{remark}
	
	\section{Design of Exogenous Input and Estimator}
	\label{Design}
	\subsection{Exogenous Input}
	In this subsection, we discuss the design of the exogenous input. First, we apply a stochastic control input $\actionU_t$ to the system that will be fully specified shortly, and solve the differential equation $\eqref{Linear SDE}$ for that. Upon this construction, in the next subsection we build a least-squares estimator $\estdynamicA$ for the dynamics matrix $\dynamicA$. Analysis of this two-fold procedure will be presented in the next section, where we establish high probability error bounds. 
	
	Let us proceed by defining
	\begin{equation}
		\label{definition of scaled sde}
		\scaledstate_t = e^{-\dynamicA t}\state_t.
	\end{equation}
	Since according to Ito calculus, the product of two differentials vanishes \cite{karatzas2014brownian}, we obtain the following for $\scaledstate_t$
	\begin{equation}
		\label{scaledlinear SDE}
		d\scaledstate_t = -e^{\dynamicA t}\dynamicA\state_tdt+e^{-\dynamicA t}d\state_t.
	\end{equation}
	Now, replace $\eqref{Linear SDE}$ in $\eqref{scaledlinear SDE}$, to get
	\begin{equation}
		\label{scaledlinear SDE2}
		d\scaledstate_t = e^{-\dynamicA t}\controlB\actionU_tdt+e^{-\dynamicA t}Cd\WeinerPP_t.
	\end{equation}
	Therefore, the fundamental theorem of calculus yields to
	\begin{equation}
		\label{scaled Linea SDE 2}
		\scaledstate_t = \scaledstate_0 +\int_{0}^{t}e^{-\dynamicA s}\controlB\actionU_sds + \int_{0}^{t}e^{-\dynamicA s}Cd\WeinerPP_s.
	\end{equation}
	Using $\scaledstate_0=\state_0$, together with \eqref{definition of scaled sde} and \eqref{scaled Linea SDE 2}, and finally by multiplying both sides by $e^{\dynamicA t}$, we obtain the solution
	\begin{equation}
		\label{scaled linear SDE 3}
		\state_t=e^{\dynamicA t}X_0	+\int_{0}^{t}e^{\dynamicA (t-s)}\controlB\actionU_sds+\int_{0}^{t}e^{\dynamicA (t-s)}Cd\WeinerP_s.
	\end{equation}
	Then, to precisely design the control input $\actionU_t$, let 
	\begin{equation}
		\label{action}
		\actionU_s=\kappa\frac{ d\WeinerControl _s}{ds},
	\end{equation}
	where $\WeinerControl_s\in \mathbb{R}^{\dimaction}$ is a $\dimaction$-dimensional Wiener process independent of the $\dimnoise$-dimensional stochastic disturbance $\WeinerPP_s$. In other words, $U_s$ is the product of a scalar $\kappa \geq 1$ and  a $\dimaction$-dimensional white noise $d\WeinerControl_s/ds$~\cite{lokka2004stochastic,sobczyk2013stochastic}. Plugging into $\eqref{scaled linear SDE 3}$, we get 
	\begin{equation}
		\label{Linear SDE3}
		\state_t=e^{\dynamicA t}X_0	+\kappa\int_{0}^{t}e^{\dynamicA (t-s)}\controlB d\WeinerControl _s+\int_{0}^{t}e^{\dynamicA (t-s)}Cd\WeinerP_s.
	\end{equation}
	So, the second term above becomes an Ito integral (similar to the third term)~\cite{lokka2004stochastic}. Additional technicalities in the above derivation as well as the definition of $\actionU_s$ in \eqref{action}, fall beyond the scope of this work and can be found in the relevant literature on generalized derivatives of Wiener processes~\cite{lokka2004stochastic,sobczyk2013stochastic}. 
	
	The two stochastic integrals on the right hand side of $\eqref{Linear SDE3}$ possess conceptually different roles in our setting because by the definition, the disturbance noise $\WeinerPP_s$ is unobservable, while we have a full knowledge of the Wiener \emph{signal} $\WeinerControl_s$ that we design and apply to the system as the control action. We leverage this knowledge in the design of the estimator in the next subsection. 
	
	\subsection{Least Squares Estimator}
	The objective is to find an estimator which minimizes the total sum of squares, i.e., the aggregated deviations of the predicted Ito differential form from the actually observed one, as elaborated below. To proceed, note that by plugging $\actionU_t$ in $\eqref{action}$ into the dynamics equation $\eqref{Linear SDE}$, we have
	\begin{equation}
		\label{formal sde}
		d\state_t = \dynamicA \state_t dt + \kappa \controlB d\WeinerControl_t+Cd\WeinerPP_t.
	\end{equation}
	Therefore, for $A \in \mathbb{R}^{\dimstate \times \dimstate}$, we define the loss function as
	\begin{equation}
		\label{loss definition}
		\mathcal{L}_T(A)=\int_{0}^{T} \left\lVert \frac{ d \state_t }{ dt } - \kappa\controlB \frac{ d \WeinerControl_t }{ dt } - A\state_t \right\rVert_2^2 dt.
	\end{equation}
	So, after solving for a matrix $A$ that minimizes $\mathcal{L}_T(A)$, the least-squares estimate of $\dynamicA$ will be 
	\begin{equation*}
		\estdynamicA= \left[\int_{0}^{T}\state_t\left(d\state_t-\kappa\controlB d\WeinerControl_t\right)^\top\right]^{\top} \left[\int_{0}^{T}\state_t\state_t^\top dt\right]^{\dagger} .
	\end{equation*}  
	A slightly more precise derivation of the above estimate according to the Riemann definition of the integral in $\eqref{loss definition}$ is available in the literature \cite{faradonbeh2023online}, \cite{ibrahimi2013three}.
	
	Later on, we show that the first matrix on the right hand side of $\eqref{estimator2}$
	is invertible for $T$ being sufficiently large. Therefore, for all such $T$, $\estdynamicA$ can be equivalently written as
	\begin{equation}
		\label{estimator2}
		\estdynamicA= \left[\int_{0}^{T}\state_t\left(d\state_t-\kappa\controlB d\WeinerControl_t\right)^\top\right]^{\top} \left[\int_{0}^{T}\state_t\state_t^\top dt\right]^{-1} .
	\end{equation}
	
	\section{Main Results}
	\label{theoretical results}
	After doing some algebra, since $\dynamicA$ satisfies \eqref{formal sde}, the error of the estimate in $\eqref{estimator2}$ can be written as follows:
	\begin{equation}
		\label{error}
		\estdynamicA-\dynamicA= C\left[\int_{0}^{T}\state_td\WeinerPP_t^\top\right]^\top\left[\int_{0}^{T}\state_t\state_t^\top dt\right]^{-1}.
	\end{equation}
	Thus, the analysis of the estimator heavily relies on the following two random matrices. First, the sample covariance matrix 
	\begin{equation}
		\label{sample covariance matrix}
		V_T = \int_{0}^{T} \state_t\state_t^\top dt,
	\end{equation}
	and second, the matrix-valued martingale
	\begin{equation}
		\label{martingale transformation}
		S_T = \int_{0}^{T}\state_t d\WeinerPP_t^\top.
	\end{equation}
	In the sequel, we establish tight bounds on $V_T$, and $S_T$ (after being self-normalized by $V_T$), leading to the following bound for $\estdynamicA-\dynamicA$.
	
	\begin{theorem}
		\label{maintheorem}
		For any $\delta > 0$, with probability at least $1-\delta$, we have
		\begin{equation}
			\label{stableerror}
			\tnorm{\estdynamicA-\dynamicA}_2^2 = \mathcal{O}\left(\frac{\dimstate\tnorm{C}_2^2\left(\log T-\log\delta\right)}{cT\kappa^2}\right),
		\end{equation}
		when $\dynamicA$ is stable, 
		\begin{equation}
			\label{marginallystableerror}
			\tnorm{\estdynamicA-\dynamicA}_2^2 = \mathcal{O}\left(\frac{\dimstate^2\tnorm{C}_2^2\left(\log T-\log\delta\right)}{cT\kappa^2}\right),
		\end{equation}
		when $\dynamicA$ is marginally stable, and whenever $\dynamicA$ is unstable,
		\begin{equation}
			\label{unstableerror}
			\tnorm{\estdynamicA-\dynamicA}_2^2 = \mathcal{O}\left(\frac{ \dimstate\tnorm{C}_2^2\left(T-\log\delta\right)}{c T \kappa^2}\right).
		\end{equation}	
	\end{theorem}
	
	Theorem $\ref{maintheorem}$ states that the estimation error scales with $\dimstate$ when $\dynamicA$ is marginally stable and with $\sqrt{\dimstate}$ when it is not. Moreover, the error scales with $\tnorm{C}_2/\kappa$, which is the inverse of the signal-to-noise ratio (SNR). Further, it shows that when the system is (marginally) stable, the estimator $\eqref{estimator2}$ has square-root consistency up to a logarithmic factor. However, when the system is unstable, a larger value for $\kappa$ is needed to ensure an estimation error of the conventional square-root decay rate; $\kappa = \sqrt{T}$ guarantees that, according to $\eqref{unstableerror}$. In general, a larger value for $\kappa$ leads to a faster decay rate of the estimation error. However, larger values for $\kappa$ leads to larger $\state_t$ which defeats the purpose \cite{dean2020sample}. Therefore, the magnitude $\kappa$ should not be excessively large.


	Next, we define the size of the largest block in the Jordan decomposition of the open-loop matrix; $\dynamicA=P^{-1}\Lambda P$. That is, $\Lambda$ is block-diagonal with blocks $\Lambda_1, \cdots, \Lambda_k$, each Jordan block $\Lambda_i$ being a matrix of size $l_i$ for the eigenvalue $\lambda_i$:
	\begin{equation*}
		\Lambda = 	\begin{bmatrix}
			\Lambda_{1} &\cdots & 0\\
			\vdots& \ddots & \vdots\\
			0&\cdots & \Lambda_{k},
		\end{bmatrix}, ~~~
		\Lambda_i = \begin{bmatrix}
			\lambda_i & 1 & 0 &\cdots & 0\\
			0 & \lambda_i &  1&\cdots  & 0\\
			\vdots & \vdots & \vdots&\ddots  & \vdots \\
			0 & 0  & 0&\cdots  & \lambda_i
		\end{bmatrix}.
	\end{equation*}
	Then, let the size of the largest Jordan block be
	\begin{equation}
		\label{Jordan size}
		l^* = \max\left\{l_i,i=1,\cdots,k\right\}.
	\end{equation}
	
	\begin{remark}
		\label{jordan remark}
		When the system is marginally stable, we can replace the factor $\dimstate^2$ in \eqref{marginallystableerror} with $l^*\dimstate$, where $l^*$ is defined in $\eqref{Jordan size}$, and in the worst case (that $\Lambda$ has only one block) is $\dimstate$.
	\end{remark}
	\section{Proof of  the Theorem and Auxiliary Lemmas}
	\label{auxiliary}
	In this section, we first present three lemmas which constitute the building blocks for the proof of Theorem \ref{maintheorem}. Then, we employ these intermediate results for completing the proof. Proofs of the following auxiliary lemmas are lengthy, involves extensive technical details, relies on further intermediate results, and so are delegated to the supplementary materials. 
	
	Lemma \ref{Self_Norm lemma} establishes high-probability upper-bounds for continuous-time matrix-valued self-normalized martingales. 
	\begin{lemma}
		\label{Self_Norm lemma}
		Let $V_T$ and $S_T$ be as in \eqref{sample covariance matrix} and \eqref{martingale transformation}, respectively. Further, define $\bar{V}_T = V_T +V$, where $V$ is an arbitrary positive definite matrix. Then, with probability at least $1-\delta$,
		\begin{equation*}
			\tnorm{\bar{V}_T^{\frac{-1}{2}}S_T}_2^2 \leq 8\log\left(\frac{12^{\dimnoise}~ \det(\bar{V}_T)^{\frac{1}{2}}}{\delta~ \det(V)^{\frac{1}{2}}}\right)\  \text{for\ all\ $T\geq 0$}.
		\end{equation*}
	\end{lemma}
	
	The next lemma provides a high probability upper-bound for the sample covariance matrix. 
	
	
	\begin{lemma}
		\label{eigmaxlemma}
		Let $l^*$ and $\kappa$ be as in $\eqref{Jordan size}$ and $\eqref{action}$ respectively. For $\delta>0$, with probability at least $1-\delta$, we have
		\begin{equation}
			\lambda_{\max}(\covmatrixV_T) = \mathcal{O}\left(\left(\dimstate\kappa^2T^{2l^*}e^{2\lambda_{\max}(\dynamicA)T}+\kappa^2T\right)\log \frac{T}{\delta}\right).
		\end{equation}
	\end{lemma}

	Note that based on this lemma, when $\dynamicA$ is unstable, the largest eigenvalue of $V_T$ grows exponentially fast with time, specifically as $\mathcal{O}(\kappa^2T^{2l^*}e^{2\lambda_{\max}(\dynamicA)T})$. However, in stable and marginally stable systems, the growth is polynomial and linear, i.e., with the rates $\mathcal{O}(\kappa^2T^{2l^*})$ and $\mathcal{O}(\kappa^2T)$, respectively.
	
	The next result provides a lower-bound for the sample covariance matrix $V_T$ in \eqref{sample covariance matrix}.
	
	\begin{lemma}
		\label{eigminlemma}
		In stable, marginally stable, and unstable systems, and for any $\delta>0$, with probability at least $1-\delta$, it holds that
		\begin{equation*}
			\lambda_{\max}(V_T^{-1}) =  \mathcal{O}\left(\frac{1}{cT\kappa^2}\right).
		\end{equation*}
	\end{lemma}
	
	\begin{figure*}[thbt!]
		\centering
		\label{figure1}
		\begin{subfigure}{0.32\textwidth}
			\centering
			\includegraphics[width=\textwidth]{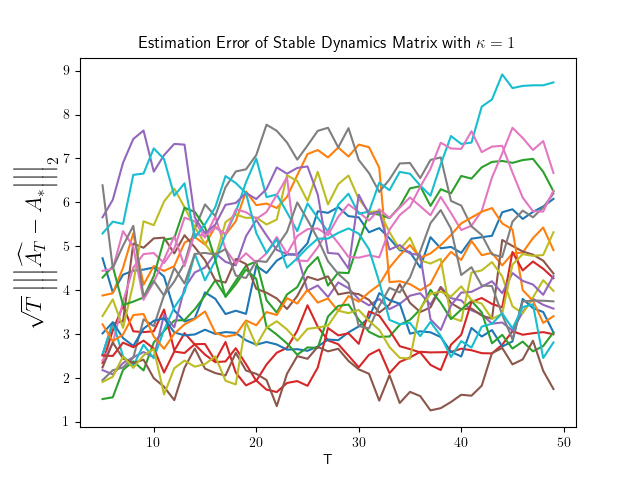}
			\caption{Estimation error amplified by a $\sqrt{T}$ factor vs $T$, for a stable system with $\kappa = 1$.}
			\label{fig:plot1}
		\end{subfigure}
		\hfill
		\begin{subfigure}{0.32\textwidth}
			\centering
			\includegraphics[width=\textwidth]{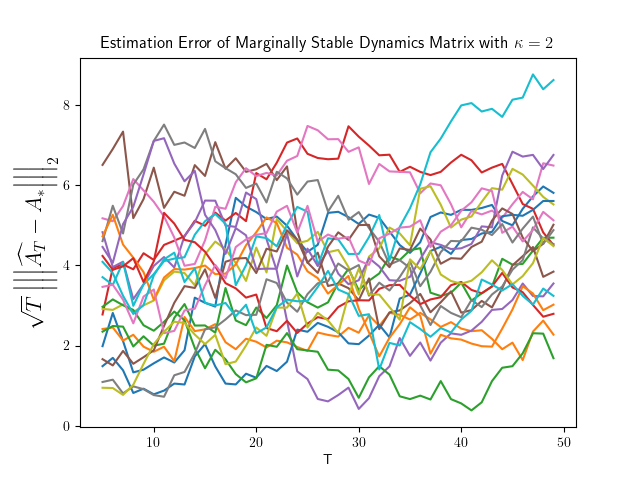}
			\caption{Estimation error amplified by $\sqrt{T}$ vs $T$, for a marginally stable system with $\kappa = 2$.}
			\label{fig:plot2}
		\end{subfigure}
		\hfill
		\begin{subfigure}{0.32\textwidth}
			\centering
			\includegraphics[width=\textwidth]{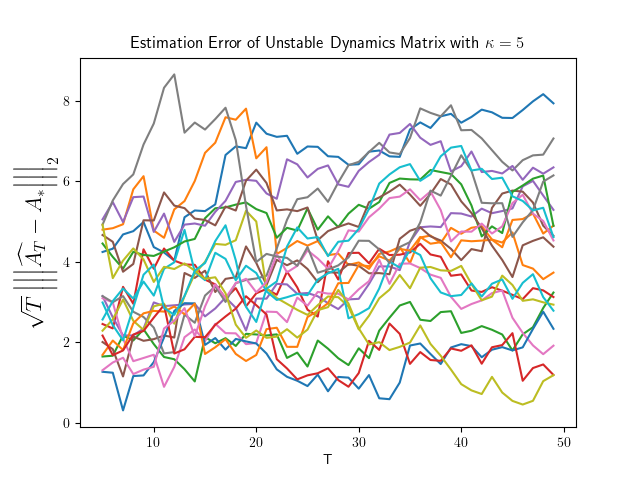}
			\caption{Estimation error amplified by a $\sqrt{T}$ factor vs $T$, for an unstable system with $\kappa = 5$.} 
			\label{fig:plot3}
		\end{subfigure}
		\caption{Multiple realizations of properly scaled estimation errors are illustrated versus the length of the state trajectory, for stable, marginally stable, and unstable systems. Further, by comparing the graphs, we see that the compromised estimation caused by system instability can be compensated by increasing $\kappa$; i.e., by employing larger random input signals.}
	\end{figure*}
	
	\subsection{Proof of Theorem~\ref{maintheorem}}
	Let $V_T$, $S_T$ be as in $\eqref{sample covariance matrix}$, $\eqref{martingale transformation}$, and define 
	\begin{equation}
		\label{definition of V}
		\bar{V}_T=V_T+\mathbb{I}_{\dimstate}.
	\end{equation}
	Clearly, for the estimation error in $\eqref{error}$, it holds that
	\begin{equation}
		\label{error22}
		\begin{split}
			\dynamicA^\top - \estdynamicA^\top &= V_T^{-1}S_TC^\top\\
			&=V_T^{-1}(V_T+\mathbb{I}_{\dimstate})^{\frac{1}{2}}\bar{V}_T^{\frac{-1}{2}}S_TC^\top\\
			&=(V_T^{-1}+V_T^{-2})^{\frac{1}{2}}\bar{V}_T^{\frac{-1}{2}}S_TC^\top.
		\end{split}
	\end{equation}
	Next, we have the following inequality
	\begin{equation}
		\label{useful inequality}
		(V_T^{-1}+V_T^{-2})^{\frac{1}{2}} \leq V_T^{\frac{-1}{2}}+V_T^{-1}.
	\end{equation}
	Note that the above is implied by
	\begin{equation*}
		\begin{split}
			(V_T^{\frac{-1}{2}}+V_T^{-1})^2=V_T^{-1}+V_T^{-2}+2V_T^{\frac{-3}{2}},
		\end{split}
	\end{equation*}
	because the last term in the latter inequality is a positive semidefinite matrix. Therefore, \eqref{error22} together with \eqref{useful inequality}, lead to
	\begin{equation}
		\label{error33}
		\begin{split}
			\tnorm{\estdynamicA-\dynamicA}_2 &\leq \tnorm{(V_T^{-1}+V_T^{-2})^{\frac{1}{2}}}_2\tnorm{\bar{V}_T^{\frac{-1}{2}}S_T}_2\tnorm{C}_2\\
			&\leq \tnorm{V_T^{\frac{-1}{2}}+V_T^{-1}}_2\tnorm{\bar{V}_T^{\frac{-1}{2}}S_T}_2\tnorm{C}_2.
		\end{split}
	\end{equation}
	So, by applying the triangle inequality, we obtain
	\begin{equation*}
		\begin{split}
			\tnorm{\estdynamicA-\dynamicA}_2	&\leq \tnorm{V_T^{\frac{-1}{2}}}_2\tnorm{\bar{V}_T^{\frac{-1}{2}}S_T}_2\tnorm{C}_2\\
			&+\tnorm{V_T^{-1}}_2\tnorm{\bar{V}_T^{\frac{-1}{2}}S_T}_2\tnorm{C}_2.
		\end{split}
	\end{equation*}
	When $T$ is large enough, according to Lemma  $\ref{eigminlemma}$, with probability at least $1-\delta$, it holds that 
	\begin{equation}
		\label{error 44}
		\tnorm{\estdynamicA-\dynamicA}_2	\leq 2\tnorm{V_T^{\frac{-1}{2}}}_2\tnorm{\bar{V}_T^{\frac{-1}{2}}S_T}_2\tnorm{C}_2.
	\end{equation}
	We bound the first term on the right hand side of $\eqref{error 44}$ using Lemma $\ref{eigminlemma}$, and leverage Lemma \ref{Self_Norm lemma} to upper-bound the second term. Thus, with probability at least $1-2\delta$, we have
	\begin{equation*}
		\begin{split}
			\tnorm{\estdynamicA-\dynamicA}_2 = & \mathcal{O}\left(\sqrt{\frac{ \tnorm{C}_2^2}{cT\kappa^2}}\sqrt{\log\frac{12^{\dimnoise}(\lambda_{\max}(V_T))^{\frac{\dimstate}{2}}}{\delta}}\right).
		\end{split}
	\end{equation*}
	Note that $\dimstate$ and $\dimnoise$ have the same order of magnitude. Thus, by expanding the logarithm, the above estimation error bound becomes
	\begin{equation}
		\label{logarithm expansion}
		\begin{split}
			\mathcal{O}\left(\sqrt{\frac{ \dimstate\tnorm{C}_2^2\left(\log\left(\lambda_{\max}(V_T)\right)-\log\delta\right)}{cT\kappa^2}}\right).
		\end{split}
	\end{equation}
	When $\dynamicA$ is stable, we have the following bound from Lemma $\ref{eigmaxlemma}$, and its subsequent discussion, with probability at least $1-\delta$:
	\begin{equation}
		\label{stableupperbound}
		\log\left(\lambda_{\max}(V_T)\right) = \mathcal{O}\left(\log T-\log\delta\right).
	\end{equation}
	Therefore, in this case, by combining $\eqref{logarithm expansion}$ and $\eqref{stableupperbound}$, we get the desired error bound in $\eqref{stableerror}$.  
	
	On the other hand, when $\dynamicA$ is marginally stable, Lemma $\eqref{eigmaxlemma}$ indicates that with probability at least $1-\delta$, we have
	\begin{equation}
		\label{marginallystableupperbound}
		\log\left(\lambda_{\max}(V_T)\right) = \mathcal{O}\left(l^*\log T-\log\delta\right),
	\end{equation}
	where $l^*$ is defined in $\eqref{Jordan size}$, and as explained in the Remark \ref{jordan remark}, can be as large as the dimension $\dimstate$. Thus, in this case, we get the rate in \eqref{marginallystableerror} by putting \eqref{logarithm expansion} and \eqref{marginallystableupperbound} together.
	
	Finally, Lemma \ref{eigmaxlemma} show that when $\dynamicA$ is unstable, with probability at least $1-\delta$, it holds that
	\begin{equation}
		\label{unstableupperbound}
		\log\left(\lambda_{\max}(V_T)\right) = \mathcal{O}\left(T-\log\delta\right),
	\end{equation}
	which together with $\eqref{logarithm expansion}$ gives the high-probability bound in $\eqref{unstableerror}$, and so completes the proof.

	\section{Numerical Illustrations}
	\label{numerical}
	Next, we perform numerical simulations to illustrate the result of Theorem \ref{maintheorem}. To that end, we use an exemplary linear dynamical system representing a nuclear reactor~\cite{kerlin2019dynamics}, whose open-loop dynamics matrix is 
	\begin{equation*}
		\dynamicA = \begin{bmatrix}
			-1 & 0 & -z \\
			2 & -2 &  0\\
			0 & 3 & -3 
		\end{bmatrix}.
	\end{equation*} 
	When $z=5$, the system matrix is stable with the largest eigenvalue real-part being $\lambda_{\max}(\dynamicA)=-0.3928$. However, $z = 10$ and $z=15$, yield marginally stable and unstable systems, with the largest eigenvalue real-parts $\lambda_{\max}(\dynamicA)=0$ and $\lambda_{\max}(\dynamicA)=0.2779$, respectively.
	
	We let the input and noise coefficients matrices $\controlB$ and $C$ both be $1/5\times\mathbb{I}_3$, where $\mathbb{I}_3$ is the $3 \times 3$ identity matrix. For each system, we run $20$ independent trajectories for $50$ seconds, starting from a random initial state $\state_0$. When the dynamics matrix $\dynamicA$ is stable, we choose $\kappa = 1$, while for  marginally stable and unstable $\dynamicA$, we select $\kappa = 2$ and $\kappa = 5$, respectively. 
	
	The estimation is performed at integer times $T=1,\dots,50$, according to $\eqref{estimator2}$. Figures \ref{fig:plot1}, \ref{fig:plot2}, and \ref{fig:plot3}, show the (properly normalized) estimation error in the three different regimes of (in)stability, versus time. That is, the horizontal axes represent $T$, while the vertical axes depict the value of $\sqrt{T}\tnorm{\estdynamicA-\dynamicA}_2$. All the $20$ trajectories are plotted in the graphs to illustrate the high probability nature of the results in Theorem \ref{maintheorem}. 

	\section{Conclusion and Future Work}
	\label{conclusions}
	We study system identification in linear dynamical systems that evolve in continuous-time according to stochastic differential equations. The setting involves only one trajectory of the state vectors and does not assume any form of stability for system dynamics, rendering guaranteed learning under instability a key technical challenge. We propose an applicable estimation method based on least-squares estimates together with randomized control inputs.

	Then, we establish performance guarantees by quantifying the decay rates of estimation error as the length of the trajectory grows. All different regimes of stability are considered so that the open-loop dynamics can be (marginally) stable or unstable. We also provide effects of the following parameters on the quality of system identification: dimensions, signal-to-noise ratios, magnitudes of stochastic control inputs, and eigenvalues and eigenspaces of the matrix governing the system dynamics.

	Importantly, this work specifies the extent to which one can address consistency issues caused by system instability, through effective and easy-to-implement designs of the control inputs. Furthermore, the technical framework we develop for theoretical analysis of the identification performance are applicable to a wide range of problems. Accordingly, the presented approach can be extended to a family of relevant problems that constitute interesting subjects for future work. That includes problems involving identification of both the input and state-transition matrices, as well as extensions to systems of slower excitations, higher order differential equations, and positive-definite controllability Grammians. 
	
	\bibliographystyle{IEEEtran}
	\bibliography{bib}

	\section{appendices}
	\label{appendix}
	\subsection{Organization of the Appendices}
	The appendices are organized as follow. Appendix \ref{complementary appendix} includes the complementary notation and additional definitions that set up the stage for the proofs of Lemmas \ref{Self_Norm lemma}, \ref{eigmaxlemma}, and \ref{eigminlemma}, which are given in the appendices \ref{proof of self normal lemma app}, \ref{proof of eigmax lemma app}, and \ref{proof of eigmin lemma app}, respectively.
	\subsection{Complementary Notation and Definitions}
	\label{complementary appendix}
	\begin{enumerate}
		\item For a matrix $A$, we define $\tnorm{A}_{\alpha\rightarrow \beta}$ indexed by $\alpha$ and $\beta$ as $\tnorm{A}_{\alpha\rightarrow \beta}= \sup_{v}\lVert Av \rVert_\beta/\lVert v \rVert_\alpha$. When $\alpha = \beta$, we simply write $\tnorm{A}_{\alpha}$. The Frobenius norm of the matrix $A$ is indicated by $\tnorm{A}_F$. For complex number $\lambda$, $\Re(\lambda)$ is the real part of $\lambda$. We adopt the convention that $\inf\{\emptyset\}=\infty$. Moreover, $\vee$ and $\wedge$ stand for the maximum and minimum respectively.
		\item For a positive semi-definite matrix $P\in \mathbb{R}^{n\times n}$, and vectors $v \in \mathbb{R}^{n}$, we define the $P$-norm of $v$ as $\lVert v\rVert_{P} = \left(v^\top P v\right)^{\frac{1}{2}}$. 
		\item Let $x_t$ be a 1-dimensional random process given by
		\begin{equation*}
			x_t = x_0 +\int_{0}^{t} \mu_s ds + \int_{0}^{t} \sigma_s d\weinerP_s,
		\end{equation*}
		where $x_0$ is a deterministic initial point, $\mu(s)$ and $\sigma(s)$ are deterministic coefficients which only depend on the time component $s$, and $\weinerP_s$ is a 1-dimensional standard Brownian motion. We denote the quadratic variation of $x_t$ by $\llbracket x\rrbracket_t$ which is defined as
		\begin{equation*}
			\llbracket x\rrbracket_t = \int_{0}^{t}\sigma_s^2ds.
		\end{equation*}
		\item Define $H=[\kappa \controlB,C]$, that is, the matrix $H$ is obtained by stacking the columns of the two matrices $\kappa\controlB$ and $C$. 
		\item Similarly, define $L = [\controlB, C]$.
		\item Lemma \ref{Self_Norm lemma} considers deviations of the stochastic integral
		\begin{equation*}
			\int_{0}^{T}\state_td\WeinerPP_t^\top,
		\end{equation*}
		when appropriately transformed by $\bar{V}_T$. However, a similar result can be shown for the stochastic integral
		\begin{equation*}
			\int_{0}^{T}\state_t d\WeinerControl_t^\top,
		\end{equation*}
		where $\WeinerControl_t$ denotes the Weiner process coming from the control action $\actionU_t$. Indeed we can combine these two random processes and get another similar result for this combined process. To make this rigorous, let $\widetilde{\WeinerPP}_t$ be the random process obtained from concatenation of $\WeinerControl_t$ and $\WeinerPP_t$. That is 
		\begin{equation*}
			\widetilde{\WeinerPP}_t = \begin{bmatrix}\WeinerControl_t   \\\WeinerPP_t\end{bmatrix}, 
		\end{equation*}
		and let us define $\widetilde{S}_T$ and $\widetilde{y}(T,\delta)$ accordingly
		\begin{equation*}
			\widetilde{S}_T = \int_{0}^{T}\state_td\widetilde{\WeinerPP}_t^\top,
		\end{equation*}
		\begin{equation*}
			\widetilde{y}(T,\delta) = \sqrt{8\log\left(\frac{12^{(\dimaction+\dimnoise)}\ \det(\bar{V}_T)^{\frac{1}{2}}}{\delta\det(V)^{\frac{1}{2}}}\right)}.
		\end{equation*}
		Note that the only difference between $y(T,\delta)$ and $\widetilde{y}(T,\delta)$, is the change in exponent of the constant $12$, which reflects the dimension effect of the concatenated process.
		With $V_T$ and $V$ defined as before, we have the following re-statement of Lemma \ref{Self_Norm lemma}: 
		For any $\delta>0$ we have
		\begin{equation*}
			\mathbb{P}\left(\tnorm{\bar{V}_T^{\frac{-1}{2}}\widetilde{S}_T}_2 < \widetilde{y}(T,\delta)\  \text{for\ all\ $T\geq 0$}\right)\geq 1-\delta.
		\end{equation*}
		\item From now on, we drop the tilde symbol from $\widetilde{\WeinerPP}_t$, $\widetilde{S}_T$, and $\widetilde{y}(T,\delta)$, and simply write them as $\WeinerPP_t$, $S_T$, and $y(T,\delta)$. Thus, throughout all the proofs, by $\WeinerPP_t$ we mean this new stacked random process, and $S_T$ and $y(T,\delta)$ have altered accordingly. 
	\end{enumerate}
	\subsection{Proof of Lemma $\ref{Self_Norm lemma}$}
	\label{proof of self normal lemma app}
	This proof is based on the follwoing propositions which are presented first, and are followed by the proof of the Lemma \ref{Self_Norm lemma}. \newline
	\begin{proposition}
		\label{self-normalization prop1}
		Let $w \in \mathcal{S}^{\dimaction+\dimnoise-1}$ be an arbitrary element of the unit sphere. Moreover, let $\theta \in \mathbb{R}^{\dimstate}$ be an arbitrary vector. Now, define
		
		\begin{equation*}
			\begin{split}
				M_t^{\theta} &= \exp\left\{\int_{0}^{t}\theta^\top\state_s d\WeinerPP_s^\top w-\frac{1}{2}\int_{0}^{t}\theta^\top\state_s\state_s\theta ds\right\},
			\end{split}
		\end{equation*}
		and let $\tau$ be a stopping time with respect to the filtration $\{\mathcal{F}_t\}$.
		Then $M_{\tau}^{\theta}$ is almost surely well-defined and is a super martingale with $\mathbb{E}(M_{\tau}^{\theta})\leq1$.\newline
	\end{proposition}
	\subsubsection{Proof of Proposition \ref{self-normalization prop1}}
	
	Let $t_2>t_1>0$. We show that
	\begin{equation*}
		\mathbb{E}\left[M_{t_2}^{\theta}|\mathcal{F}_{t_1}\right]=M_{t_1}^{\theta}.
	\end{equation*}
	
	Choose $\epsilon$ such that we have the following discretization of $[0,t_2]=0,\epsilon,\dots,N_1\epsilon=t_1,\dots,N_2\epsilon = t_2$, and let $\WeinerPP_i \sim \mathbb{N}(0,\epsilon \mathbb{I}_{\dimaction+\dimnoise}), i=1,\dots,N_2$ be the corresponding discretization of the Weiner process. Note that $w^\top \WeinerPP_i\sim \mathbb{N}(0,\epsilon)$ since $w$ is a unit vector. For $i=0,\dots,N_2-1$ define
	\begin{equation*}
		D_i^{\theta} = \exp\left(\theta^\top \state_i \WeinerPP_{i+1}^\top w-\frac{\epsilon}{2}\theta^\top \state_i\state_i^\top\theta\right).
	\end{equation*}
	By known properties of the moment generating function of a normal variable, for any $i$ we have
	\begin{equation*}
		\mathbb{E}\left[D_i^{\theta}|\mathcal{F}_{(i-1)\epsilon}\right]= 1.
	\end{equation*}
	Moreover, from the tower property of conditional expectation, we can write
	\begin{equation*}
		\mathbb{E}\left[D_{i+1}^{\theta}|\mathcal{F}_{(i-1)\epsilon}\right]=\mathbb{E}\left[\mathbb{E}\left[D_{i+1}^{\theta}|\mathcal{F}_{(i)\epsilon}\right]|\mathcal{F}_{(i-1)\epsilon}\right]= 1,
	\end{equation*}
	and so on and so forth. Therefore, we obtain the following martingale equality
	\begin{equation*}
		\begin{split}
			\mathbb{E}\left[M_{t_2}|\mathcal{F}_{t_1}\right]
			=\lim_{\epsilon \rightarrow0}\mathbb{E}\left[M_{t_1}^{\theta}D_{N_1+1}^{\theta}\dots D_{N_2}^{\theta}|\mathcal{F}_{t_1}\right]
			=M_{t_1}^{\theta} \lim_{\epsilon \rightarrow0}\mathbb{E}\left[D_{N_1+1}^{\theta}\dots D_{N_2}^{\theta}|\mathcal{F}_{N_1\epsilon}\right]= M_{t_1}^{\theta}
		\end{split}.
	\end{equation*}
	
	Now, suppose $\tau$ is a stopping time with respect to the filtration $\left\{\mathcal{F}_t\right\}$, and let $Y_t^{\theta}=M_{\min (\tau, t)}^{\theta}$ be the stopped version of $M_t^{\theta}$. It is well-known that the stopped version of a martingale is also martingale. Therefore, we can write
	\begin{equation*}
		\mathbb{E}\left[Y_t^{\theta}\right] = \mathbb{E}\left[Y_0^{\theta}\right]=1.
	\end{equation*}
	By the convergence theorem for non-negative martingales, $\lim\limits_{t\rightarrow\infty}Y_t^{\theta}$ almost surely exists. This means that $M_{\tau}^{\theta}$ is almost surely well defined. Now, by application of Fatou's Lemma, we get
	\begin{equation*}
		\begin{split}
			\mathbb{E}\left[M_{\tau}^{\theta}\right]
			=\mathbb{E}\left[\liminf_{t\rightarrow \infty}Y_t^{\theta}\right]
			\leq \liminf_{t\rightarrow \infty}\mathbb{E}\left[Y_t^{\theta}\right]\leq 1		
		\end{split}.\newline
	\end{equation*} 
	\begin{proposition}
		\label{self-normalization prop2}
		Let $w \in \mathcal{S}^{\dimaction+\dimnoise-1}$ and $\tau$ be a stopping time with respect to the filtration $\{\mathcal{F}_t\}$. Moreover, let $V$ be any positive definite $\dimstate\times \dimstate$ matrix and $\bar{V}_T=V_T+V$, then with probability at least $1-\delta$ we have
		
		\begin{equation*}
			\lVert S_{\tau}w \rVert_{\bar{V}_{\tau}^{-1}} \leq 2\log\left(\frac{\det\left(\bar{V}_{\tau}\right)^{\frac{1}{2}}}{\delta\det\left(V\right)^{\frac{1}{2}}}\right).
		\end{equation*}
		\newline
		\newline
	\end{proposition}
	\subsubsection{Proof of Proposition \ref{self-normalization prop2}}
	
	Let $\Theta \sim \mathbb{N}(0_{\dimstate},V^{-1})$ be a normal vector with covariance matrix $V^{-1}$. Define
	\begin{equation*}
		M_t = \int_{\mathbb{R}^{\dimstate}}M_t^{\theta}f(\theta)d(\theta),
	\end{equation*} 
	where $f$ is the density function of $\Theta$. Note that we have $\mathbb{E}[M_{\tau}]=\mathbb{E}[\mathbb{E}[M_{\tau}^{\Theta}|\Theta]]\leq 1$. Therefore, we can write
	\begin{equation*}
		M_t =(2\pi)^{\frac{-\dimstate}{2}} \det(V)^{\frac{1}{2}} \int_{\mathbb{R}^{\dimstate}}\exp\left\{\theta^\top S_tw-\frac{1}{2}\theta^\top(V+V_t)\theta\right\}d\theta.
	\end{equation*}
	Let $\bar{V}_t=V_t+V$. By completing the square inside the exponential, we get $M_t$ as the following expression
	\begin{equation*}
		\begin{split}
			&\left(\frac{\det(V)}{\det(\bar{V}_t)}\right)^{\frac{1}{2}}\exp\left\{\frac{1}{2}w^\top S_t^\top(\bar{V}_t)^{-1}S_tw\right\}
			\int_{\mathbb{R}^{\dimstate}}\frac{\det(\bar{V}_t)^{\frac{1}{2}}}{(2\pi)^{\frac{\dimstate}{2}}}\exp\left\{-\frac{1}{2}\left[(\theta-(\bar{V}_t)^{-1}S_tw)^\top(\bar{V}_t)(\theta-(\bar{V}_t)^{-1}S_tw)\right]\right\}d\theta\\
			=&\left(\frac{\det(V)}{\det(\bar{V}_t)}\right)^{\frac{1}{2}}\exp\left\{\frac{1}{2}w^\top S_t^\top(\bar{V}_t)^{-1}S_tw\right\}.
		\end{split} 
	\end{equation*}
	With some simple algebra, the following equality holds
	\begin{equation*}
		\begin{split}
			&\ \mathbb{P}\left(w^\top S_{\tau}^\top\bar{V}_{\tau}^{-1}S_{\tau}w>2\log\left(\frac{\det(\bar{V}_{\tau})^{\frac{1}{2}}}{\delta \det(V)^{\frac{1}{2}}}\right)\right)\\
			=&\ \mathbb{P}\left(\exp\left(\frac{1}{2}w^\top S_{\tau}^\top(\bar{V}_{\tau})^{-1}S_{\tau}w\right)\left(\frac{\det(V)}{\det(\bar{V}_{\tau})}\right)^{\frac{1}{2}}>\frac{1}{\delta}\right),
		\end{split}
	\end{equation*}
	which by application of the Markov inequality and the Proposition \ref{self-normalization prop1}, can be bounded by
	\begin{equation*}
		\begin{split}
			&\delta\mathbb{E}\left[\exp\left(\frac{1}{2}w^\top S_{\tau}^\top(\bar{V}_{\tau})^{-1}S_{\tau}w\right)\left(\frac{\det(V)}{\det(\bar{V}_{\tau})}\right)^{\frac{1}{2}}\right]\\
			=&\ \delta\mathbb{E}\left[M_{\tau}\right]\\
			\leq&\ \delta.
		\end{split}
	\end{equation*}

	\begin{proposition}
		\label{no. of points in epsilon net}
		Let $O^{m\times m'}, m>m'$ be the set of matrices with orthonormal columns. Then for every $\epsilon \in (0,1)$ there exist an $\epsilon-net$, $\mathcal{N}_{\epsilon} \subset O^{m\times m'} $, such that $|\mathcal{N}_{\epsilon}|\leq (\frac{6\sqrt{m'}}{\epsilon})^{mm'}$ and for every $V \in O^{m\times m'} $ there is a $V' \in \mathcal{N}_{\epsilon}$ such that $\tnorm{V-V'}_F \leq \epsilon$.\newline
		
	\end{proposition}
	The proof is available in the literature \cite{du2020few}.
	\begin{remark}
		Note that when $m'=1$ the Frobenius norm in the above proposition would be the same as the usual $l_2$ norm. In such case, the $\epsilon$-net is to cover the unit sphere in the $m$-dimensional space. 
	\end{remark}
	The following proposition is also adopted from the existing literature \cite{vershynin2018high}.
	\begin{proposition}
		\label{vershynin}
		Let $M$ be a random matrix and $\epsilon \leq 1$ arbitrary. Then we have
		\begin{equation*}
			\mathbb{P}\left(\tnorm{M}_2>z\right) \leq \mathbb{P} \left(\max_{w \in \mathcal{N}_{\epsilon}}\lVert Mw \rVert > (1-\epsilon)z\right).
		\end{equation*}\newline
	\end{proposition}
	Now, we can present the proof of Lemma $\ref{Self_Norm lemma}$.\newline
	\subsubsection{Proof of Lemma \ref{Self_Norm lemma}}
	Let $\epsilon = \frac{1}{2}$, $\mathcal{N}_{\frac{1}{2}}$ be the corresponding $\epsilon-net$, and $w \in \mathcal{N}_{\frac{1}{2}}$ an arbitrary element of the $\epsilon-net$ in the $(\dimstate+\dimnoise)$-dimensional space. Based on proposition $\ref{no. of points in epsilon net}$, $|\mathcal{N}_{\frac{1}{2}}|\leq 12^{\dimaction+\dimnoise}$. Now define
	\begin{equation*}
		B_t(\delta) =\left\{\omega_0 \in \Omega: w^\top S_t^\top\bar{V}_t^{-1}S_tw > 2\log\left(\frac{12^{\dimaction+\dimnoise}\det(\bar{V}_t)^{\frac{1}{2}}}{\delta\det(V)^{\frac{1}{2}}}\right)\right\},
	\end{equation*}
	and 
	\begin{equation*}
		\tau(\omega_0) = \min\left\{t\geq 0:\omega_0 \in B_t(\delta)\right\}.
	\end{equation*}
	Note that $\bigcup_{t\geq0} B_t(\delta)=\{\omega_0:\tau(\omega_0)<\infty\}=\left\{w^\top S_{\tau}^\top\bar{V}_{\tau}S_{\tau}w>2\log\left(\frac{12^{\dimaction+\dimnoise}\ \det(\bar{V}_{\tau})^{\frac{1}{2}}}{\delta\det(V)^{\frac{1}{2}}}\right)\right\}$. So we obtain
	\begin{equation*}
		\begin{split}
			&\mathbb{P}\left(\bigcup_{t\geq0} B_t(\delta)\right)\\
			=&\mathbb{P}\left(\tau <\infty\right)\\
			=&\mathbb{P}\left(w^\top S_{\tau}^\top\bar{V}_{\tau}S_{\tau}w>2\log\left(\frac{12^{\dimaction+\dimnoise}\ \det(\bar{V}_{\tau})^{\frac{1}{2}}}{\delta\det(V)^{\frac{1}{2}}}\right)\right)\\
			\leq&\frac{\delta}{12^{\dimaction+\dimnoise}}.
		\end{split}
	\end{equation*}
	The last inequality comes from proposition $\ref{self-normalization prop2}$. Now using proposition $\ref{vershynin}$ we have the desired result
	\begin{equation*}
		\begin{split}
			&\mathbb{P}\left(\tnorm{\bar{V}_t^{\frac{-1}{2}}S_t}_2 > y(t,\delta) \text{ for some } t\geq 0\right)\\
			\leq &\mathbb{P}\left(\max_{w\in \mathcal{N}_{\frac{1}{2}}}w^\top S_t^\top\bar{V}_t^{-1}S_tw>\frac{y(t,\delta)^2}{4}\text{ for some }t\geq0\right),\newline
		\end{split}
	\end{equation*}
	which by union bounds over all points of the epsilon-net, can be bounded as 
	\begin{equation*}
		\sum_{w \in \mathcal{N}_{\frac{1}{2}}}\mathbb{P}\left(w^\top S_t^\top\bar{V}_t^{-1}S_tw>\frac{y(t,\delta)^2}{4} \text{ for some }t\geq 0\right)
		\leq \delta.\newline
	\end{equation*}
	\subsection{Proof of Lemma $\ref{eigmaxlemma}$}
	\label{proof of eigmax lemma app}
	This proof constructs upon the following theorem which is available in the literature \cite{revuz2013continuous}, and states that every 1-dimensional continuous local martingale vanishing at zero can be represented as a standard Brownian motion with a changed clock. We use this theorem together with a finite-time version of the law of iterated logarithm \cite{howard2021time} to construct the upper bound. For the sake of completeness, we state these two theorems in terms of Propositions $\ref{dubins-shwarz}$ and $\ref{finite time law of iterated logarithm2}$ and further discuss their implications. 
	\begin{proposition}
		\label{dubins-shwarz}
		Let $x_t$ be a 1-dimensional $\mathcal{F}_t$-measurable continuous local martingale vanishing at zero, with $\lim_{t \rightarrow \infty}\llbracket x\rrbracket_t = \infty, $ almost surely. For each $0\leq s<\infty$, define the stopping time
		\begin{equation*}
			T(s)=\left\{\inf{t\geq0;\llbracket x\rrbracket_t>s}\right\}.
		\end{equation*}
		Then the time changed process $\weinerP_s=x_{T_s}$ is a standard Brownian motion with respect to the filtration $\mathcal{F}_{T_s}$, and we have
		\begin{equation*}
			x_t = \weinerP_{\llbracket x\rrbracket_t};\ 0\leq t<\infty.
		\end{equation*}
	\end{proposition}
	
	The interpretation is that, $x$ can be considered as a standard Brownian motion $\weinerP$, that runs with a different clock. This means, we replace the original clock $t$ with the changed clock $\llbracket x\rrbracket_t$. Therefore we can translate known properties of the Brownian motion, for other local martingales. One such property is the following finite-time law of iterated logarithm from the literature \cite{howard2021time}. It provides, high probability finite-time uniform bounds for the displacement of a standard Brownian motion.  
	\begin{proposition}
		\label{finite time law of iterated logarithm2}
		Let $\weinerP_t$ be the standard 1-dimensional Brownian motion. Then for any $\eta>1$ and $s>1$
		\begin{equation}
			\label{finite time law of iterated logarithm}
			\mathbb{P}\left(\exists t\in(0,\infty):\weinerP_t\geq\frac{\eta^{\frac{1}{4}}+\eta^{\frac{-1}{4}}}{\sqrt{2}}\sqrt{(1\vee t)\left(s\log\log(\eta(1\vee t))+\log\frac{\zeta(s)}{\delta\log^s\eta}\right)}\right)\leq \delta,
		\end{equation}
		where $\zeta(s)$ is the Riemann zeta function.  \newline
	\end{proposition}
	
	As an example, Suppose $x_t$ is a 1-dimensional linear SDE satisfying $dx_t=\eigenvalue x_tdt+hd\weinerP_t^1,\ x_0=0$ where $\eigenvalue$ is a complex valued number with corresponding real part $\realeigenvalue$, $h\in \mathbb{R}$ is a positive number, and $\weinerP^1$ is a standard $1$-dimensional Brownian motion. Note that $x_t$ can be equivalently shown as $x_t=\int_{0}^{t}e^{\eigenvalue(t-s)}hd\weinerP_s^1$. The quadratic variation of $x_t$ is
	\begin{equation*}
		\begin{split}
			\llbracket x\rrbracket_t &= \int_{0}^{t}\lVert e^{\eigenvalue(t-s)}h\rVert^2ds\\
			&=h^2\int_{0}^{t}e^{2\realeigenvalue(t-s)}ds.
		\end{split}
	\end{equation*}
	It is easy to check that when $\realeigenvalue>0$, $x_t$ meets the condition of the Proposition \ref{dubins-shwarz}, and therefore can be represented as a Brownian motion with a changed clock. That is, $x_t = \weinerP_{\alpha_{\realeigenvalue, h}(t)},\ 0\leq t<\infty$, where $\alpha_{\realeigenvalue, h}(t)=\frac{h^2}{2\realeigenvalue}(e^{2\realeigenvalue t}-1)$ is the new clock. Note that when $\realeigenvalue<0$ the condition $\lim_{t\rightarrow\infty}\llbracket x\rrbracket_t=\infty$, is not met. However, Proposition \ref{dubins-shwarz} is still applicable under an enlargement of the filtered probability space. A rigorous argument is available in the literature \cite{revuz2013continuous}. 
	Now, we consider the three cases of unstable, marginally stable, and stable dynamics matrix separately. 
	\subsubsection{Unstable dynamics, $\Re\left(\lambda_1(\dynamicA)\right)>0$}
	First, suppose we have an integral of the form

	\begin{equation}
		\label{one dim xx}
		x_t=\int_{0}^{t}e^{\eigenvalue(t-s)}hd\weinerP_s,
	\end{equation}
	for a complex valued $\eigenvalue$, with positive real part $\realeigenvalue$, and a positive real number $h$. Define
	\begin{equation*}
		\alpha_{\realeigenvalue, h}(t)= \frac{h^2}{2\realeigenvalue}(e^{2\realeigenvalue t}-1), 
	\end{equation*}
	and 
	\begin{equation}
		\label{definition of f unstable}
		f_{\realeigenvalue,h}(t,\delta) = h\sqrt{\frac{2}{\realeigenvalue}}e^{\realeigenvalue t}\sqrt{2\log(2\realeigenvalue t+1+\log\frac{h^2}{2\realeigenvalue})+\log\frac{4}{\delta}}.
	\end{equation}
	By letting $s=2\ and\ \eta = e$ in $\eqref{finite time law of iterated logarithm}$ and application of Proposition $\ref{dubins-shwarz}$, one can easily check the following high probability bound
	\begin{equation}
		\label{high prob bound}
		\mathbb{P}\left(|x_t|\leq f_{\realeigenvalue,h}(t,\delta),\text{for all t }\geq 0\right)\geq 1-\delta.
	\end{equation}
	Now, we can upper bound  $\lVert \state_t\rVert$ in the general setting. Suppose $\dynamicA$ has the Jordan decomposition $\dynamicA=P^{-1}\Lambda P$, where $\Lambda$ is a block diagonal matrix
	\begin{equation*}
		\Lambda = 	\begin{bmatrix}
			\Lambda_{1} &\cdots & 0\\
			\vdots& \ddots & \vdots\\
			0&\cdots & \Lambda_{k}
		\end{bmatrix},~~~
		\Lambda_i = \begin{bmatrix}
			\eigenvalue_i & 1 & 0 &\cdots & 0\\
			0 & \eigenvalue_i &  1&\cdots  & 0\\
			\vdots & \vdots & \vdots&\ddots  & \vdots \\
			0 & 0  & 0&\cdots  & \eigenvalue_i
		\end{bmatrix},
	\end{equation*}
	and each Jordan block $\Lambda_i$ has $\eigenvalue_i$ on the main diagonal, $1$ right of diagonal and zero elsewhere. Also, without loss of generality, we assume that the eigenvalues in the above Jordan decomposition, are ordered based on their real parts. That is $\Re{\lambda_1}\geq\Re{\lambda_2}\geq\dots\geq \Re{\lambda_{k}}$
	 Note that $\state_t$ can be written as
	\begin{equation}
		\label{solution of the linear SDE}
		\state_t = P^{-1}e^{\Lambda t}PX_0+P^{-1}\int_{0}^{t}e^{\Lambda(t-s)}PHd\WeinerPP_s.
	\end{equation}
	Now, we bound each term on the right hand side of the above expression. First, note that for each Jordan block, we have
	\begin{equation}
		\label{matrix exponential}
		e^{\Lambda_i(t-s)}=e^{\eigenvalue_i(t-s)}\begin{bmatrix}
			1 & (t-s) & \frac{(t-s)^2}{2!} &\cdots & \frac{(t-s)^{l_i-1}}{(l_i-1)!}\\
			0 & 1 &  (t-s)&\cdots  & \frac{(t-s)^{l_i-2}}{(l_i-2)!}\\
			\vdots & \vdots & \vdots&\ddots  & \vdots \\
			0 & 0  & 0&\cdots  & 1
		\end{bmatrix}.
	\end{equation}
	Therefore, we can expand the stochastic term on the right hand side of $\eqref{solution of the linear SDE}$ as follow. Let $c_1^\top,\cdots c_\dimstate^\top$ be the $\dimstate$-dimensional row vectors that constitute the rows of $P$. Moreover, suppose $\weinerP^i, i=1,\dots,\dimaction+\dimnoise$ are the 1-dimensional Brownian components of the noise vector $\WeinerPP$. Therefore, $PHd\WeinerPP_s$ is a vector in $\mathbb{R}^{\dimstate}$ the $j^{th}$ element of which is given by 
	\begin{equation*}
		\langle c_j^\top H,d\WeinerPP_s \rangle = \sum_{i=1}^{{\dimaction+\dimnoise}}(c_j^\top H)_id\weinerP_s^i, 
	\end{equation*}
	where $(c_j^\top H)_i$ is the $i^{th}$ element of the row vector $c_j^\top H$ and $\weinerP^i$ the $i^{th}$ component of the Brownian motion. Thus, with reference to the exponential blocks $\eqref{matrix exponential}$ we define the vector $v$ as
	\begin{equation}
		\label{vee definition}
		v = \int_{0}^{t}e^{\Lambda(t-s)}PHd\WeinerP_s=\begin{bmatrix} v_1  \\ \vdots \\v_{\dimstate}\end{bmatrix},
	\end{equation}
	where 
	\begin{equation*}
		v_1 = \sum_{i=1}^{{\dimaction+\dimnoise}}\int_{0}^{t}e^{\eigenvalue_1(t-s)}(c_1^\top H)_id\weinerP_s^i+\dots+\sum_{i=1}^{\dimaction+\dimnoise}\int_{0}^{t}\frac{(t-s)^{l_1-1}}{(l_1-1)!}e^{\eigenvalue_1(t-s)}(c_{l_1}^\top H)_id\weinerP_s^i,
	\end{equation*}
	and 
	\begin{equation*}
		v_2 = \sum_{i=1}^{{\dimaction+\dimnoise}}\int_{0}^{t}e^{\eigenvalue_1(t-s)}(c_2^\top H)_id\weinerP_s^i+\dots+\sum_{i=1}^{\dimaction+\dimnoise}\int_{0}^{t}\frac{(t-s)^{l_1-2}}{(l_1-2)!}e^{\eigenvalue_1(t-s)}(c_{l_1}^\top H)_id\weinerP_s^i,
	\end{equation*}
	and so on and so forth. Therefore, we can write the following upper bound for the magnitude of the second vector on the right hand side of $\eqref{solution of the linear SDE}$
	\begin{equation}
		\label{long vector2}
		\lVert  P^{-1}v \rVert_2\leq\tnorm{P^{-1}}_{\infty \rightarrow2}\lVert v \rVert_{\infty}.
	\end{equation}
	Note that all the entries of the matrix $PH$, have magnitude at most as large as $p^*$, defined as  $p^*=\tnorm{H}_{\infty}\tnorm{P}_{\infty}$. Therefore, for all $t\geq 1$, we have with probability at least $1-\delta$
	\begin{equation}
		\label{vector2}
		\begin{split}
			\lVert  P^{-1} v\rVert_2 &\leq \tnorm{P^{-1}}_{\infty \rightarrow2}\left({\dimaction+\dimnoise}\right)f_{\realeigenvalue_1,p^*}\left(t,\frac{\delta}{(\dimaction+\dimnoise)\dimstate l^*}\right)\left(1+t+\dots+\frac{t^{(l^*-1)}}{(l^*-1)!}\right)\\
			&\leq \tnorm{P^{-1}}_{\infty \rightarrow2}({\dimaction+\dimnoise})et^{l^*-1}f_{\realeigenvalue_1,p^*}\left(t,\frac{\delta}{(\dimaction+\dimnoise)\dimstate l^*}\right)\\
			&\leq\sqrt{\frac{ 2}{\Re\left(\lambda_1\right)}}\beta \dimstate^{\frac{1}{2}} e\kappa
			({\dimaction+\dimnoise})e^{\Re\left(\lambda_1\right)t}t^{l^*-1}\sqrt{2\log(2\Re\left(\lambda_1\right)t+1+\log\frac{{p^*}^2}{2\Re\left(\lambda_1\right)})+\log\frac{4\dimstate l^*(\dimaction+\dimnoise)}{\delta}}.
		\end{split}
	\end{equation}
	where 
	\begin{equation}
		\label{definition of beta}
		\beta = \tnorm{P^{-1}}_{\infty}\tnorm{P}_{\infty}\tnorm{L}_{\infty},~L=[B,C].
	\end{equation}
	The first inequality in $\eqref{vector2}$ is based on the following reasoning. The quadratic variation of integrals of the form $\eqref{one dim xx}$ which constitute the components of $v$ is increasing in both $h$ and $\realeigenvalue$. Moreover, the uniform upper bound given in $\eqref{finite time law of iterated logarithm}$ is increasing in $t$. Therefore, we can for all $i,j,l$, uniformly bound the integrals of the form $\int_{0}^{t}e^{\eigenvalue_l(t-s)}(c_j^\top H)_id\weinerP_s^i$ with the largest upper bound given by $f_{\realeigenvalue_1,p^*}$. The second inequality in $\eqref{vector2}$ is because of the fact that 
	\begin{equation}
		\label{useful fact}
		1+\frac{1}{2}+\dots+\frac{1}{(l^*-1)!} \leq e,
	\end{equation}
	and the last inequality in $\eqref{vector2}$ is from the fact that $\tnorm{P^{-1}}_{\infty\rightarrow 2}\leq  \sqrt{\dimstate}\tnorm{P^{-1}}_{\infty}$.
	Now, we bound the first term on the right hand side of $\eqref{solution of the linear SDE}$. Note that
	\begin{equation}
		\label{deterministic bound1}
		\lVert e^{\dynamicA t}X_0 \rVert \leq \tnorm{P^{-1}}_{\infty\rightarrow2}\tnorm{e^{\Lambda t}}_{\infty}\tnorm{P}_{\infty}\lVert X_0 \rVert_{\infty}.
	\end{equation}
	Moreover, $e^{\Lambda t}$ is composed of the blocks of the form
	\begin{equation*}
		e^{\Lambda_it}=e^{\eigenvalue_it}\begin{bmatrix}
			1 & t & \frac{t^2}{2!} &\cdots & \frac{t^{l_i-1}}{(l_i-1)!}\\
			0 & 1 &  t&\cdots  & \frac{t^{l_i-2}}{(l_i-2)!}\\
			\vdots & \vdots & \vdots&\ddots  & \vdots \\
			0 & 0  & 0&\cdots  & 1
		\end{bmatrix}.
	\end{equation*}
	Therefore,  
	\begin{equation*}
		\tnorm{e^{\Lambda t}}_{\infty} \leq e^{\realeigenvalue_1 t}\left(1+\dots+\frac{t^{l^*-1}}{(l^*-1)!}\right)\leq et^{l^*-1}e^{\Re\left(\lambda_1\right) t},
	\end{equation*}
	where the last inequality comes from $\eqref{useful fact}$. Thus, from $\eqref{deterministic bound1}$ we have
	\begin{equation*}
		\lVert e^{\dynamicA t}X_0 \rVert \leq \tnorm{P^{-1}}_{\infty\rightarrow2}\tnorm{P}_{\infty}\lVert X_0 \rVert_{\infty}et^{l^*-1}e^{\Re\left(\lambda_1\right) t}.
	\end{equation*}
	Hence, we can write
	\begin{equation}
		\label{new deterministic bound}
		\lVert e^{\dynamicA t}X_0 \rVert \leq \tnorm{P^{-1}}_{\infty}\tnorm{P}_{\infty}\lVert X_0 \rVert_{\infty}\dimstate^{\frac{1}{2}} et^{l^*-1}e^{\Re\left(\lambda_1\right) t},
	\end{equation}
	Note that when $\sqrt{\frac{4\log2\Re\left(\lambda_1\right)t}{\Re\left(\lambda_1\right)}}>\frac{\lVert X_0\rVert_{\infty}}{\tnorm{L}_{\infty}}$ the upper bound in $\eqref{vector2}$ is larger than the upper bound in $\eqref{new deterministic bound}$. Therefore, putting the bound in $\eqref{vector2}$ back in $\eqref{solution of the linear SDE}$, we get the following result
	\begin{equation}
		\label{norm of the system state2}
		\lVert\state_t\rVert_2 \leq 2\sqrt{\frac{ 2}{\Re\left(\lambda_1\right)}}\beta \dimstate^{\frac{1}{2}} e\kappa
		({\dimaction+\dimnoise})e^{\Re\left(\lambda_1\right)t}t^{l^*-1}\sqrt{2\log(2\Re\left(\lambda_1\right)t+1+\log\frac{{p^*}^2}{2\Re\left(\lambda_1\right)})+\log\frac{4\dimstate l^*(\dimaction+\dimnoise)}{\delta}}.
	\end{equation}
	Thus, we have established the following upper bound for the largest eigen-value of the covariance matrix
	\begin{equation}
		\label{largest eigen-value of the covariance matrix2.}
		\lambda_{\max}(\covmatrixV_t) \leq \frac{8}{\Re\left(\lambda_1\right)}\beta^2\dimstate e^2\kappa^2
		({\dimaction+\dimnoise})e^{2\Re\left(\lambda_1\right)t}t^{2l^*-1}\left[2\log(2\Re\left(\lambda_1\right)t+1+\log\frac{{p^*}^2}{2\Re\left(\lambda_1\right)})+\log\frac{4\dimstate l^*(\dimaction+\dimnoise)}{\delta}\right].
	\end{equation}
	Moreover, define
	\begin{equation}
		\label{C u t and delta}
		C_u(t,\delta)=\frac{8}{\Re\left(\lambda_1\right)}\beta^2\dimstate e^2\kappa^2
		({\dimaction+\dimnoise})\left[2\log(2\Re\left(\lambda_1\right)t+1+\log\frac{{p^*}^2}{2\Re\left(\lambda_1\right)})+\log\frac{4\dimstate l^*(\dimaction+\dimnoise)}{\delta}\right].
	\end{equation}
	We need $C_u(t,\delta)$ in proof of Lemma $\ref{eigminlemma}$.\newline
	
	\subsubsection{Marginally stable dynamics, $\Re\left(\lambda_1(\dynamicA)\right)=0$}
	The proof in the marginally case, is exactly the same. First, we bound the second term on the right hand side of $\eqref{solution of the linear SDE}$. Note that in this case, all the independent $1$-dimensional random components $\int_{0}^{t}e^{\eigenvalue_l(t-s)}(c_j^\top H)_id\weinerP_s^i$ which are components of $v$ $\eqref{vee definition}$, can be uniformly bounded with probability at least with $\frac{\delta}{\dimstate l^*(\dimaction+\dimnoise)}$ $f_{p^*}(t,\frac{\delta}{\dimstate l^*(\dimaction+\dimnoise)})$, where we define
	
	\begin{equation}
		\label{definition of f marginally stable }
		f_{h}(t,\delta) = 
		2ht^{\frac{1}{2}}\sqrt{2\log(\log t+2\log h+1)+\log\frac{4}{\delta}}, \realeigenvalue>0.
	\end{equation}
	Therefore, the second term on the right hand side of $\eqref{solution of the linear SDE}$ can be uniformly bounded with probability at least $1-\delta$ as follow
	\begin{equation}
		\label{vector3}
		\begin{split}
			\lVert  P^{-1} v\rVert_2 &\leq \tnorm{P^{-1}}_{\infty \rightarrow2}({\dimaction+\dimnoise})f_{p^*}\left(t,\frac{\delta}{(\dimaction+\dimnoise)\dimstate l^*}\right)\left(1+t+\dots+\frac{t^{(l^*-1)}}{(l^*-1)!}\right)\\
			&\leq \tnorm{P^{-1}}_{\infty \rightarrow2}({\dimaction+\dimnoise})et^{l^*-\frac{1}{2}}f_{p^*}\left(t,\frac{\delta}{(\dimaction+\dimnoise)\dimstate l^*}\right)\\
			&\leq2\beta \dimstate^{\frac{1}{2}} e\kappa
			({\dimaction+\dimnoise})t^{l^*-\frac{1}{2}}\sqrt{2\log(\log t+2\log p^*+1)+\log\frac{4\dimstate l^*(\dimaction+\dimnoise)}{\delta}}.
		\end{split}
	\end{equation}
	Similarly, the first term on the right hand side of $\eqref{solution of the linear SDE}$ can be bounded as follow
	\begin{equation*}
		\lVert e^{\dynamicA t}X_0 \rVert \leq \tnorm{P^{-1}}_{\infty\rightarrow2}\tnorm{P}_{\infty}\lVert X_0 \rVert_{\infty}et^{l^*-1}.
	\end{equation*}
	Hence, we can write
	\begin{equation*}
		\lVert e^{\dynamicA t}X_0 \rVert \leq \tnorm{P^{-1}}_{\infty}\tnorm{P}_{\infty}\lVert X_0 \rVert_{\infty}\dimstate^{\frac{1}{2}} et^{l^*-1},
	\end{equation*}
	which can be equivalently written as
	\begin{equation}
		\label{new deterministic bound margin}
		\lVert e^{\dynamicA t}X_0 \rVert \leq \frac{\beta}{\tnorm{L}_{\infty}}\lVert X_0 \rVert_{\infty}\dimstate^{\frac{1}{2}} et^{l^*-1}.
	\end{equation}
	Note that when 
	\begin{equation*}
		t^{\frac{1}{2}}>\frac{\lVert X_0\rVert_{\infty}}{\tnorm{L}_{\infty}},
	\end{equation*}
	the upper bound in $\eqref{vector3}$ is at least twice larger than the upper bound in $\eqref{new deterministic bound margin}$. Therefore, putting the bound in $\eqref{vector3}$ back in $\eqref{solution of the linear SDE}$, we get the following result
	\begin{equation}
		\label{norm of the system state}
		\lVert\state_t\rVert_2 \leq 3\beta \dimstate^{\frac{1}{2}} e\kappa
		({\dimaction+\dimnoise})t^{l^*-\frac{1}{2}}\sqrt{2\log(\log t+2\log p^*+1)+\log\frac{4\dimstate l^*(\dimaction+\dimnoise)}{\delta}}\newline.
	\end{equation}
	Hence, with probability at least $1-\delta$ we have the following upper bound for the largest eigen-value of the covariance matrix
	\begin{equation}
		\label{largest eigen-value of the covariance matrix.}
		\lambda_{max}(\covmatrixV_t) \leq 9\beta^2 \dimstate e^2\kappa^2
		({\dimaction+\dimnoise})^2t^{2l^*}\left[2\log(\log t+2\log p^*+1)+\log\frac{4\dimstate l^*(\dimaction+\dimnoise)}{\delta}\right].
	\end{equation}
	Next, we make the following definition which is used in the proof of Lemma $\ref{eigminlemma}$.
	\begin{equation}
		\label{C s in marginally stable}
		C_s(t, \delta)= 9\beta^2 \dimstate e^2\kappa^2
		({\dimaction+\dimnoise})^2\left[2\log(\log t+2\log p^*+1)+\log\frac{4\dimstate l^*(\dimaction+\dimnoise)}{\delta}\right].
	\end{equation}

	\subsubsection{Stable dynamics matrix, $\Re\left(\lambda_1(\dynamicA)\right)<0$}
	First, suppose we have an integral of the form

	\begin{equation}
		\label{one dim xx stable}
		x_t=\int_{0}^{t}e^{\eigenvalue(t-s)}(t-s)^lhd\weinerP_s,
	\end{equation}
	for a complex valued $\eigenvalue$ with negative real part $\realeigenvalue$, a one dimensional Weiner process $\weinerP$, and a positive real number $h$ and a non-negative integer $l$. Then, $x_t$ has a quadratic variation $\alpha_{\realeigenvalue, h}(t)$ given by the following integral 
	
	\begin{equation*}
		\alpha_{\realeigenvalue, h}(t)= h^2\int_{0}^{t} e^{2\realeigenvalue s}s^{2l} ds
	\end{equation*}
	Note that $\alpha_{\realeigenvalue, h}(t)$ is increasing in both $\realeigenvalue$, and $h$ .With multiple applications of integration by parts, we get the following closed form solution for $\alpha_{\realeigenvalue, h}(t)$
	\begin{equation*}
		\alpha_{\realeigenvalue, h}(t)= h^2e^{2\realeigenvalue t}\left[\sum_{i=0}^{2l}\frac{(t^{2l-i})2l!}{(-2\realeigenvalue)^{i+1}(2l-i)!}\right].
	\end{equation*}
	Now, we bound both terms on the right hand side of $\eqref{solution of the linear SDE}$. To bound the second term, first let $t_0$ be large enough such that $\alpha_{\realeigenvalue_1, p^*}(t)<1$. Moreover, let 
	\begin{equation*}
		l_0 = \max_{{0\leq s\leq t_0}}\{\alpha_{\realeigenvalue_1, p^*}(s)\} \vee 1
	\end{equation*}
	Note that since $\alpha_{\realeigenvalue_1, p^*}$ is continuous, and we take the maximum on a compact set, then $l_0$ is bounded. Let
	\begin{equation}
		\label{definition of f stable }
		f_{h}(t,\delta) = 
		\sqrt{2l^0\left(2\log\log l^0 + \log \frac{4}{\delta}\right)},
	\end{equation}
	which is a constant with respect to $t$. Hence, the second term on the right hand side of $\eqref{solution of the linear SDE}$ can be uniformly bounded with probability at least $1-\delta$ as follow
	
	\begin{equation}
		\label{vector4}
		\begin{split}
			\lVert  P^{-1} v\rVert_2 &\leq \tnorm{P^{-1}}_{\infty \rightarrow2}({\dimaction+\dimnoise})f_{p^*}\left(t,\frac{\delta}{(\dimaction+\dimnoise)\dimstate l^*}\right)\left(1+\dots+\frac{1}{(l^*-1)!}\right)\\
			&\leq \tnorm{P^{-1}}_{\infty \rightarrow2}({\dimaction+\dimnoise})e\sqrt{2l^0\left(2\log\log l^0 + \log \frac{4\dimstate l^*(\dimaction+\dimnoise)}{\delta}\right)}\\
			&\leq\beta \dimstate^{\frac{1}{2}} e\kappa
			({\dimaction+\dimnoise})\sqrt{2l^0\left(2\log\log l^0 + \log \frac{4\dimstate l^*(\dimaction+\dimnoise)}{\delta}\right)}.
		\end{split}
	\end{equation}
	Similar to the unstable case, we can bound the deterministic term on the right hand side of $\eqref{solution of the linear SDE}$ as follow

	\begin{equation*}
		\lVert e^{\dynamicA t}X_0 \rVert \leq \tnorm{P^{-1}}_{\infty\rightarrow2}\tnorm{P}_{\infty}\lVert X_0 \rVert_{\infty}et^{l^*-1}e^{\Re\left(\lambda_1\right) t}.
	\end{equation*}
	Hence, we can write
	\begin{equation}
		\label{new deterministic bound stable}
		\lVert e^{\dynamicA t}X_0 \rVert \leq \tnorm{P^{-1}}_{\infty}\tnorm{P}_{\infty}\lVert X_0 \rVert_{\infty}\dimstate^{\frac{1}{2}} et^{l^*-1}e^{\Re\left(\lambda_1\right) t},
	\end{equation}
	Let $t$ be large enough such that the bound on the right hand side of $\eqref{new deterministic bound stable}$ is smaller than the bound on the right hand side of $\eqref{vector4}$. Therefore, putting the bound in $\eqref{vector4}$ back in $\eqref{solution of the linear SDE}$, we get the following result
	\begin{equation}
		\label{norm of the system state2 stable}
		\lVert\state_t\rVert_2 \leq 2\beta \dimstate^{\frac{1}{2}} e\kappa
		({\dimaction+\dimnoise})\sqrt{2l^0\left(2\log\log l^0 + \log \frac{4\dimstate l^*(\dimaction+\dimnoise)}{\delta}\right)}.
	\end{equation}
	Thus, we have the following upper bound for the largest eigen-value of the covariance matrix
	\begin{equation}
		\label{largest eigen-value of the covariance matrix2 stable}
		\lambda_{\max}(\covmatrixV_t) \leq 4\beta^2\dimstate e^2\kappa^2
		({\dimaction+\dimnoise})^2t\left[2l^0\left(2\log\log l^0 + \log \frac{4\dimstate l^*(\dimaction+\dimnoise)}{\delta}\right)\right].
	\end{equation}
	Moreover, define
	\begin{equation}
		\label{D s delta}
		D_s(\delta)=4\beta^2\dimstate e^2\kappa^2
		({\dimaction+\dimnoise})^2\left[2l^0\left(2\log\log l^0 + \log \frac{4\dimstate l^*(\dimaction+\dimnoise)}{\delta}\right)\right].
	\end{equation}
	We need $D_s(\delta)$ in proof of Lemma $\ref{eigminlemma}$.
	This completes the proof of Lemma $\eqref{eigmaxlemma}$.
	\subsection{Proof of the Lemma 3}
	\label{proof of eigmin lemma app}
	Let $Y_t = \state_t\state_t^\top$ and use Ito's formula to obtain
	\begin{equation*}
		\begin{split}
			dY_t &= d\state_t\state_t^\top + \state_td\state_t^\top +d\state_td\state_t^\top\\
			&=\dynamicA \state_t\state_t^\top dt+Hd\WeinerPP_t\state_t^\top+\state_t\state_t^\top \dynamicA^\top dt+\state_td\WeinerPP_t^\top H^\top+HH^\top dt.
		\end{split}
	\end{equation*}
	This differential equation, can be equivalently written as the following integral equation
	\begin{equation*}
		\begin{split}
			Y_T-Y_0 &= \dynamicA\Big(\int_{0}^{T}\state_t\state_t^\top dt\Big)+\Big(\int_{0}^{T}\state_t\state_t^\top dt\Big)\dynamicA^\top\\
			&+H\Big(\int_{0}^{T}d\WeinerPP_t\state_t^\top \Big)+\Big(\int_{0}^{T}\state_td\WeinerPP_t^\top\Big)H^\top+\int_{0}^{T}HH^\top dt
		\end{split}.
	\end{equation*}
	We can write this as 
	\begin{equation}
		\label{lyapunov equation1}
		\dynamicA V_T+V_T\dynamicA^\top+HS_T^\top+S_TH^\top+THH^\top+Y_0-Y_T=0.
	\end{equation}
	Let $\bar{V}_T=V_T+\mathbb{I}_\dimstate$ and $u \in \mathcal{S}^{\dimstate-1}$ be arbitrary. In equation $\eqref{lyapunov equation1}$, multiply $u^\top$ and $u$ from left and right to get
	\begin{equation*}
		u^\top \dynamicA V_Tu+u^\top V_T\dynamicA^\top u +u^\top HS_T^\top\bar{V}_T^{\frac{-1}{2}}\bar{V}_T^{\frac{1}{2}}u+u^\top \bar{V}_T^{\frac{1}{2}}\bar{V}_T^{\frac{-1}{2}}S_TH^\top u+Tu^\top HH^\top u+u^\top Y_0u-u^\top Y_Tu=0.
	\end{equation*}
	By simply writing the first four terms on the left hand side of the previous equation in the inner product notation, we obtain
	\begin{equation*}
		2\langle V_Tu,\dynamicA^\top u\rangle + 2\langle \bar{V}_T^{\frac{1}{2}}u,\bar{V}_T^{\frac{-1}{2}}S_TH^\top u\rangle +Tu^\top HH^\top u+u^\top Y_0u-u^\top Y_Tu=0.
	\end{equation*}
	Now, by application of the Cauchy-Schwarz inequality for inner products, we can write
	\begin{equation*}
		-2\lVert V_Tu\rVert\lVert \dynamicA^\top u\rVert-2\lVert\bar{V}_T^{\frac{1}{2}}u\rVert\lVert \bar{V}_T^{\frac{-1}{2}}S_TH^\top u\rVert +Tu^\top HH^\top u-u^\top Y_Tu\leq0.
	\end{equation*}
	Using the upper bound for the norm of $\bar{V}_T^{\frac{-1}{2}}S_T$ that was obtained in Lemma $\ref{Self_Norm lemma}$, with probability at least $1-\delta$ the following holds
	\begin{equation*}
		Tu^\top HH^\top u\leq 2\lVert V_Tu\rVert\lVert \dynamicA^\top u\rVert+2\lVert\bar{V}_T^{\frac{1}{2}}u\rVert\sqrt{8\log(\frac{12^{\dimaction+\dimnoise}\ \det(\bar{V}_T)^{\frac{1}{2}}\det(\mathbb{I}_{\dimstate})^{\frac{-1}{2}}}{\delta})}\tnorm{H}_2+u^\top Y_Tu.
	\end{equation*}
	Therefore, we have the following inequality with probability at least $1-\delta$
	\begin{equation}
		\label{fundamental inequality}
		Tc\kappa^2\leq 2\lVert V_Tu\rVert\lVert \dynamicA^\top u\rVert+2\kappa\lVert\bar{V}_T^{\frac{1}{2}}u\rVert\sqrt{8\log\left(\frac{12^{\dimaction+\dimnoise}\ \det(\bar{V}_T)^{\frac{1}{2}}}{\delta}\right)}\tnorm{L}_2+u^\top Y_Tu.
	\end{equation}
	Therefore, with probability $1-\delta$, at least one of the three following events must hold true
	\begin{equation*}
		\begin{split}
			\mathcal{B}_1 = \left\{\lVert V_Tu\rVert\geq\frac{1}{6\tnorm{\dynamicA}_2}\left(Tc\kappa^2\right)\right\}
		\end{split}
	\end{equation*}
	or,
	\begin{equation*}
		\begin{split}
			\mathcal{B}_2 = \left\{\kappa\lVert\bar{V}_T^{\frac{1}{2}}u\rVert\tnorm{L}_2\sqrt{8\log\left(\frac{12^{\dimaction+\dimnoise}\ \det(\bar{V}_T)^{\frac{1}{2}}}{\delta}\right)}\geq\frac{1}{6}\left(Tc\kappa^2\right)\right\}
		\end{split}
	\end{equation*}
	or,
	\begin{equation*}
		\begin{split}
			\mathcal{B}_3 = \left\{u^\top Y_Tu\geq\frac{1}{3}\left(Tc\kappa^2\right)\right\}
		\end{split}.
	\end{equation*}
	Moreover, define $\mathcal{E}_1$, and $\mathcal{E}_2$ as follow
	\begin{equation*}
		\mathcal{E}_1=\mathcal{B}_1\cup \mathcal{B}_3\ \ \ \text{and} \ \ \ \mathcal{E}_2=\mathcal{B}_2\cup \mathcal{B}_3.
	\end{equation*}
	It is clear that with probability $1-\delta$ at least one of $\mathcal{E}_1$ or $\mathcal{E}_2$ must hold. We show that under both $\mathcal{E}_1$, and $\mathcal{E}_2$ the minimum eigenvalue of the covariance matrix has the desired growth rate. First, we discuss the intuition behind defining $\mathcal{E}_1$ and $\mathcal{E}_2$. Note that
	\begin{equation*}
		u^\top V_T u =\int_{0}^{T}u^\top Y_t u dt.
	\end{equation*}
	Since $Y_t$ is path-wise continuous for all $t$, this can be written in the differential form
	\begin{equation*}
		d(u^\top V_t u)|_{t=T} = u^\top Y_T u dt.
	\end{equation*}
	Therefore, what $\mathcal{E}_1$ means is that, at each fixed time $T$, either $u^\top V_T u$ or its derivative are greater than a multiple of $Tc\kappa^2$. As we show in the sequel, this implies that there is $T_0$ such that for all $T\geq T_0$, $u^\top V_T u$ is greater than a multiple of $Tc\kappa^2$. $\mathcal{E}_2$ has a similar interpretation although with a different constant factor. 
	First, suppose that $\mathcal{E}_1$ holds. This means that with probability at least $1-\delta$, the following event holds true 
	\begin{equation*}
		\frac{1}{T\kappa^2}u^\top V_Tu\geq \frac{1}{6\tnorm{\dynamicA}_2} c\ \text{or}\ \frac{(u^\top \state_T)^2}{T\kappa^2}\geq \frac{1}{3} c.
	\end{equation*}
	Define
	\begin{equation*}
		\tau_1 = \inf\left\{ T\geq1, \ s.t.\ \frac{1}{T\kappa^2}u^\top V_Tu\geq \left\{\frac{1}{6}\wedge\frac{1}{6\tnorm{\dynamicA}_2}\right\} c\right\}.
	\end{equation*}
	Note that $\tau_1\leq 2\ $ almost surely under $\mathcal{E}_1$, because otherwise, for all $1\leq t\leq 2$ we have $\frac{1}{t\kappa^2}u^\top V_tu< \frac{1}{6\tnorm{\dynamicA}_2} c$ and therefore, for all such $t$, $\frac{(u^\top X_t)^2}{t\kappa^2}\geq \frac{1}{3} c$, thus,
	\begin{equation*}
		\frac{1}{2\kappa^2}u^\top V_2u\geq\frac{1}{2}\int_{1}^{2}\frac{(u^\top X_t)^2}{\kappa^2}dt\geq\frac{1}{2}\int_{1}^{2}\frac{(u^\top X_t)^2}{t\kappa^2}dt\geq \frac{1}{6} c
	\end{equation*} 
	which is a contradiction. This means that almost surely under $\mathcal{E}_1$, there is $T_0\leq 2$ such that 
	\begin{equation*}
		\frac{1}{T_0c\kappa^2} u^\top V_{T_0} u \geq  \left\{\frac{1}{6}\wedge\frac{1}{6\tnorm{\dynamicA}_2}\right\}.
	\end{equation*}
	Next, we show that for all $T\geq T_0$ the same relation holds. Define
	\begin{equation*}
		\tau_2 = \inf\left\{ T\geq \tau_1,\ s.t.\ \frac{1}{T\kappa^2}u^\top V_Tu<\left\{\frac{1}{6}\wedge\frac{1}{6\tnorm{\dynamicA}_2}\right\} c\right\}.
	\end{equation*}
	Since under $\mathcal{E}_1$, $\tau_1\leq 2\ $ almost surely, $\tau_2$ is well defined. Next, we prove $\tau_2=\infty\ $ almost surely. To see that, suppose $\tau_2=T_l$ for some finite $T_l$. This means that $u^\top V_{T_l}u<\frac{1}{6\tnorm{\dynamicA}_2} cT_l\kappa^2$ and therefore, $(u^\top X_{T_l})^2>\frac{1}{3} c T_l\kappa^2$. Note that the process $X_t$ is continuous at all points $t$ and therefore $V_t$ is differentiable. Therefore, we can write
	\begin{equation*}
		d(u^\top V_tu)|_{t=T_l} = (u^\top X_{T_l})^2dt.
	\end{equation*}
	Hence
	\begin{equation*}
		\begin{split}
			u^\top V_{T_l}u &= u^\top V_{T_l^-}u+d(u^\top V_{T_l}u)|_{t=T_l}\\
			&=u^\top V_{T_l^-}u+(u^\top X_{T_l})^2dt\\
			&\geq \left\{\frac{1}{6}\wedge\frac{1}{6\tnorm{\dynamicA}_2}\right\} cT_l^-\kappa^2+\frac{1}{3} cT_l\kappa^2 dt\\
			&\geq \left\{\frac{1}{6}\wedge\frac{1}{6\tnorm{\dynamicA}_2}\right\} c T_l\kappa^2
		\end{split}
	\end{equation*}
	which is a contradiction. Thus far, we have shown that under $\mathcal{E}_1$, $\lambda_{\min}(V_T)$ has the desired growth rate. Next, we show that the same order holds under $\mathcal{E}_2$.\newline
	We note that when the dynamics matrix $\dynamicA$ is stable or marginally stable, that is when $\Re\left(\lambda_1\right)<0$ or $\Re\left(\lambda_1\right)=0$ then for large enough $T$, we have $\mathcal{B}_2 \subseteq \mathcal{B}_1$ and therefore, $\mathcal{E}_2 \subseteq \mathcal{E}_1$. To see this, suppose $\mathcal{B}_2$ holds true, therefore
	\begin{equation*}
		\kappa\lVert\bar{V}_T^{\frac{1}{2}}u\rVert\tnorm{L}_2\sqrt{8\log\left(\frac{12^{\dimaction+\dimnoise}\ \det(\bar{V}_T)^{\frac{1}{2}}}{\delta}\right)}\geq\frac{1}{6}\left(Tc\kappa^2\right),
	\end{equation*}
	or equivalently,
	\begin{equation}
		\label{bound of bar v}
		\lVert\bar{V}_T^{\frac{1}{2}}u\rVert\geq\sqrt{\frac{Tc\kappa^2}{6\tnorm{\dynamicA}_2}}\sqrt{\frac{Tc\kappa^2\tnorm{\dynamicA}_2}{48\kappa^2\tnorm{L}_2^2\log\left(\frac{12^{\dimaction+\dimnoise}\ \det(\bar{V}_T)^{\frac{1}{2}}}{\delta}\right)}}.
	\end{equation}
	Note that, in stable and unstable dynamics $\lambda_{\max}(V_T)$ grows poly-logarithmic. That is, with probability $1-\delta$
	\begin{equation}
		\label{boundeigmaxstable}
		\lambda_{\max}(V_T)\leq C_s(T,\delta)T^{2l^*}e^{2\Re\left(\lambda_1\right)T}+D_s(T,\delta)T
	\end{equation}
	where $C_s(T, \delta)$ is defined in the unstable and marginally stable cases as in $\eqref{C u t and delta}$, and $\eqref{C s in marginally stable}$ respectively, and $D_s(T,\delta)$ is defined as in $\eqref{D s delta}$. Thus, when $T$ is large enough such that
	\begin{equation*}
		\frac{Tc\kappa^2\tnorm{\dynamicA}_2}{48\kappa^2\tnorm{L}_2^2\log\left(\frac{12^{\dimaction+\dimnoise}\ \left(C_s(T,\delta)T^{2l^*}e^{2\Re\left(\lambda_1\right)T}+D_s(T,\delta)T\right)^{\frac{\dimstate}{2}}}{\delta}\right)}>2,
	\end{equation*}
	then by squaring $\eqref{bound of bar v}$, we have 
	\begin{equation}
		\label{latest bound}
		\Vert\bar{V}_Tu\rVert \geq 2\frac{Tc\kappa^2}{6\tnorm{\dynamicA}_2}.
	\end{equation}
	Note that $\Vert\bar{V}_Tu\rVert\leq \Vert V_Tu\rVert+1$, hence when $Tc\kappa^2>6\tnorm{\dynamicA}_2$, from $\eqref{latest bound}$, we get
	\begin{equation}
		\lVert V_Tu\rVert \geq \frac{Tc\kappa^2}{6\tnorm{\dynamicA}_2}.
	\end{equation}
	Thus far, we have shown that when the dynamics is stable or marginally stable, $\mathcal{E}_2$ implies $\mathcal{E}_1$ with probability at least $1-\delta$. Finally, we show that when the dynamics is unstable, that is $\Re\left(\lambda_1\right)>0$, then under $\mathcal{E}_2$, the smallest eigen-value of the covariance matrix has the desired growth rate with probability at least $1-\delta$. Under the event $\mathcal{E}_2$ we have
	\begin{equation}
		\kappa\lVert\bar{V}_T^{\frac{1}{2}}u\rVert\tnorm{L}_2\sqrt{8\log\left(\frac{12^{\dimaction+\dimnoise}\ \det(\bar{V}_T)^{\frac{1}{2}}}{\delta}\right)}\geq\frac{1}{6}\left(Tc\kappa^2\right)\text{ or }u^\top Y_Tu\geq \frac{1}{3}Tc\kappa^2
	\end{equation}
	and therefore, under $\mathcal{E}_2$ we have with probability $1-\delta$
	\begin{equation}
		\label{expand the logarithm}
		\kappa\lVert\bar{V}_T^{\frac{1}{2}}u\rVert\tnorm{L}_2\sqrt{8\log\left(\frac{12^{\dimaction+\dimnoise}\ \left(C_u(T,\delta)T^{2l^*-1}e^{2\Re\left(\lambda_1\right)T}+1\right)^{\frac{\dimstate}{2}}}{\delta}\right)}\geq\frac{1}{6}\left(Tc\kappa^2\right)\text{ or }u^\top Y_Tu\geq \frac{1}{3}Tc\kappa^2
	\end{equation}
	where $C_u(T,\delta)$ is defined in $\eqref{C u t and delta}$. By expanding the logarithm in $\eqref{expand the logarithm}$ we obtain
	\begin{equation*}
		\lVert\bar{V}_T^{\frac{1}{2}}u\rVert\kappa\tnorm{L}_2\sqrt{8\left[(\dimaction+\dimnoise)\log12+\Re\left(\lambda_1\right)\dimstate T+\log\frac{\left(C_u(T,\delta)T^{2l^*-1}+1\right)^{\frac{\dimstate}{2}}}{\delta}\right]}\geq\frac{1}{6}\left(Tc\kappa^2\right) \text{or}\ \frac{(u^\top X_T)^2}{T\kappa^2}\geq \frac{1}{3} c, 
	\end{equation*}
	or equivalently
	\begin{equation*}
		\lVert\bar{V}_T^{\frac{1}{2}}u\rVert\tnorm{L}_2\sqrt{\frac{8}{T}\left[(\dimaction+\dimnoise)\log12+\Re\left(\lambda_1\right)\dimstate T+\log\frac{\left(C_u(T,\delta)T^{2l^*-1}+1\right)^{\frac{\dimstate}{2}}}{\delta}\right]}\geq\frac{1}{6}\left(\sqrt{T}c\kappa\right) \text{or}\ \frac{(u^\top X_T)^2}{T\kappa^2}>\frac{1}{3} c.
	\end{equation*}
	Now, let $T_0$ be large enough such that
	\begin{equation*}
		\frac{1}{T_0}\left[(\dimaction+\dimnoise)\log12+\log\frac{\left(C_u(T_0,\delta)T_0^{2l^*}+1\right)^{\frac{\dimstate}{2}}}{\delta}\right]<\Re\left(\lambda_1\right)\dimstate.
	\end{equation*}
	Therefore, for all $T\geq T_0$, with probability at least $1-2\delta$, the following event must hold true
	\begin{equation*}
		\lVert\bar{V}_T^{\frac{1}{2}}u\rVert\tnorm{L}_2\sqrt{16\Re\left(\lambda_1\right)\dimstate}\geq\frac{1}{6}\left(\sqrt{T}c\kappa\right) \text{or}\ \frac{(u^\top X_T)^2}{T\kappa^2}\geq \frac{1}{3} c.
	\end{equation*}
	Equivalently, for all such $T$, the following event must hold true
	\begin{equation*}
		\frac{1}{T\kappa^2}u^\top \bar{V}_Tu\geq\left(\frac{c}{24\tnorm{L}_2\sqrt{\Re\left(\lambda_1\right)\dimstate}}\right)^2 \text{or}\ \frac{(u^\top X_T)^2}{T\kappa^2}\geq\frac{1}{3} c.
	\end{equation*}
	The remaining of the proof, is very much similar to the proof for $\mathcal{E}_1$.
	Define
	\begin{equation*}
		\tau_3 = \inf\left\{ T\geq T_0\ s.t.\ \frac{1}{T\kappa^2}u^\top \bar{V}_Tu\geq\left\{\frac{1}{6}\wedge\frac{c}{\left(24\tnorm{L}_2\sqrt{\Re\left(\lambda_1\right)\dimstate}\right)^2}\right\} c\right\}.
	\end{equation*}
	Note that $\tau_3\leq 2T_0\ $ almost surely, because otherwise, for all $T_0\leq t\leq 2T_0$ we have $\frac{1}{t\kappa^2}u^\top \bar{V}_tu< \left(c/\left(24\tnorm{L}_2\sqrt{\Re\left(\lambda_1\right)\dimstate}\right)\right)^2$ and therefore, for all such $t$, $\frac{(u^\top X_t)^2}{t\kappa^2}\geq\frac{1}{3} c$, hence,
	\begin{equation*}
		\frac{1}{2T_0\kappa^2}u^\top\bar{V}_{2T_0}u\geq\frac{1}{2}\int_{T_0}^{T_0+1}\frac{(u^\top X_t)^2}{T_0\kappa^2}dt\geq \frac{1}{2}\int_{T_0}^{T_0+1}\frac{(u^\top X_t)^2}{t\kappa^2}dt\geq\frac{1}{6} c
	\end{equation*} 
	which is a contradiction. Now define
	\begin{equation*}
		\tau_4 = \inf\left\{ T\geq \tau_3,\ s.t.\ \frac{1}{T\kappa^2}u^\top \bar{V}_Tu<\left\{\frac{1}{6}\wedge\frac{c}{\left(24\tnorm{L}_2\sqrt{\Re\left(\lambda_1\right)\dimstate}\right)^2}\right\} c\right\}.
	\end{equation*}
	Since $\tau_3\leq 2T_0\ $ almost surely, $\tau_4$ is well defined. Next, we prove $\tau_4=\infty\ $ almost surely To see that, suppose $\tau_4=T_l$ for some finite $T_l$. This means that $u^\top\bar{V}_{T_l}u<\frac{c}{\left(24\tnorm{L}_2\sqrt{\Re\left(\lambda_1\right)\dimstate}\right)^2} cT_l\kappa^2$ and therefore, $(u^\top X_{T_l})^2\geq\frac{1}{3} c T_l\kappa^2$. Note that the process $X_t$ is continuous at all points $t$ and therefore $\bar{V}_t$ is differentiable. Therefore, we can write
	\begin{equation*}
		d(u^\top\bar{V}_tu)|_{t=T_l} = (u^\top X_{T_l})^2dt.
	\end{equation*}
	Hence we have,
	\begin{equation*}
		\begin{split}
			u^\top\bar{V}_{T_l}u &= u^\top\bar{V}_{T_l^-}u+d(u^\top\bar{V}_{T_l}u)|_{t=T_l}\\
			&=u^\top\bar{V}_{T_l^-}u+(u^\top X_{T_l})^2dt\\
			&\geq\frac{c}{\left(24\tnorm{L}_2\sqrt{\Re\left(\lambda_1\right)\dimstate}\right)^2} cT_l^-\kappa^2+\frac{1}{3} cT_l\kappa^2 dt\\
			&\geq\left\{\frac{1}{6}\wedge\frac{c}{\left(24\tnorm{L}_2\sqrt{\Re\left(\lambda_1\right)\dimstate}\right)^2}\right\} c T_l\kappa^2.
		\end{split}
	\end{equation*}
	Therefore, by letting $C_4=\min\left\{\frac{1}{6},\frac{1}{6\tnorm{A}_2},\frac{c}{\left(24\tnorm{L}_2\sqrt{\Re\left(\lambda_1\right)\dimstate}\right)^2}\right\}$, we have shown that for large enough $T$
	\begin{equation*}
		\lambda_{\min}(V_T) \geq C_4Tc\kappa^2,
	\end{equation*}
	which completes the proof.

\end{document}